\documentclass[lettersize,journal]{IEEEtran}
\usepackage{amsmath,amsfonts}
\usepackage{algorithmic}
\usepackage{algorithm}
\usepackage{setspace}
\usepackage{array}
\usepackage{textcomp}
\usepackage{stfloats}
\usepackage{url}
\usepackage{verbatim}
\usepackage{graphicx}
\usepackage{subfigure}
\usepackage{siunitx}
\usepackage{subcaption}
\usepackage{cite}
\usepackage{booktabs}
\usepackage{multirow}
\usepackage{hyperref}
\hypersetup{hidelinks,
	colorlinks=true,
	allcolors=black,
	pdfstartview=Fit,
	breaklinks=true}

\def\BibTeX{{\rm B\kern-.05em{\sc i\kern-.025em b}\kern-.08em
    T\kern-.1667em\lower.7ex\hbox{E}\kern-.125emX}}
\usepackage{balance}

\usepackage{threeparttable}
\newcommand{\boldres}[1]{\textbf{{#1}}}

\usepackage{makecell}

\usepackage{tabularx}

\begin{document}
\title{RED-F: Reconstruction-Elimination based Dual-stream Contrastive Forecasting for Multivariate Time Series Anomaly Prediction}
\author{PengYu~Chen,~Xiaohou~Shi,~Yuan~Chang,~Yan~Sun,~and~Sajal~K. Das,~\IEEEmembership{Fellow,~IEEE}
\thanks{
This work was supported by the National Natural Science Foundation of China under Grant 62272052. Yan Sun is the corresponding author.}
\thanks{
Pengyu Chen and Yan Sun are with the School of Computer Science (National Pilot Software Engineering School), Beijing University of Posts and Telecommunications, Beijing 100876, China (e-mail: penychen@bupt.edu.cn; sunyan@bupt.edu.cn).

Xiaohou Shi and Yuan Chang are engineers of China Telecom Research Institute Beijing, China (e-mail: shixh6@chinatelecom.cn;
changy8@chinatelecom.cn).

Sajal K. Das is with the Department of Computer Science, Missouri University
of Science and Technology, Rolla, MO 65409 USA (e-mail: sdas@mst.edu).}
}

\markboth{Journal of \LaTeX\ Class Files,~Vol.~18, No.~9, September~2020}%
{How to Use the IEEEtran \LaTeX \ Templates}

\maketitle

\begin{abstract}
Anomaly prediction (AP) in multivariate time series (MTS) is crucial to ensure system dependability. Existing methods either focus solely on whether an anomaly is imminent without providing precise predictions for the future anomaly, or performing predictions directly on historical data, which is easily drowned out by the normal patterns. To address the challenges in AP task, we propose RED-F, a novel framework comprised of the Reconstruction-Elimination Model (REM) and the Dual-stream Contrastive Forecasting Model (DFM). We utilize REM to construct a baseline of normal patterns from historical data, providing a foundation for subsequent predictions of anomalies. Then DFM simultaneously predicts both the constructed normal pattern and the current window, employing a contrastive forecast that transforms the difficult AP task into a simpler, more robust task of relative trajectory comparison by computing the divergence between these two predictions. To enable the forecasting model to generate a prediction not easily obscured by normal patterns, we propose a Multi-Series Prediction (MSP) training objective to enhance its sensitivity to the current window. Extensive experiments on multiple real-world datasets demonstrate the superior capability of RED-F in anomaly prediction tasks. Our code is available at \href{http://github.com/PenyChen/RED-F}{http://github.com/PenyChen/RED-F}.

\end{abstract}


\section{Introduction}
\IEEEPARstart{T}{he} increasing prevalence of the Internet of Things (IoT) and advanced sensing technologies necessitates reliable and secure analysis of complex multivariate time series (MTS), which is crucial for the dependability and security of modern Cyber-Physical Systems (CPS) such as industrial monitoring and critical healthcare\cite{li2023survey}. Consequently, anomaly detection (AD) technology, which focuses on identifying anomalies post-occurrence, has been widely studied\cite{xu2022anomaly,yang2023dcdetector}. However, AD cannot meet the timeliness requirements of preventive maintenance in high-reliability scenarios, thereby leading to irreversible losses. Motivated by the limitation, researchers have trended toward anomaly prediction (AP), a paradigm designed to forecast future abnormal events from historical datasets\cite{zhao2024fcm,hu2024multirc}. However, anomalies are inherently rare and extensively obscured within normal data, rendering data labeling costly and impractical. Therefore, we focus on the challenging task of unsupervised time series anomaly prediction.

\begin{figure*}[t]
    \centering
    \subfigure[SMD]{\includegraphics[width=0.48\linewidth]{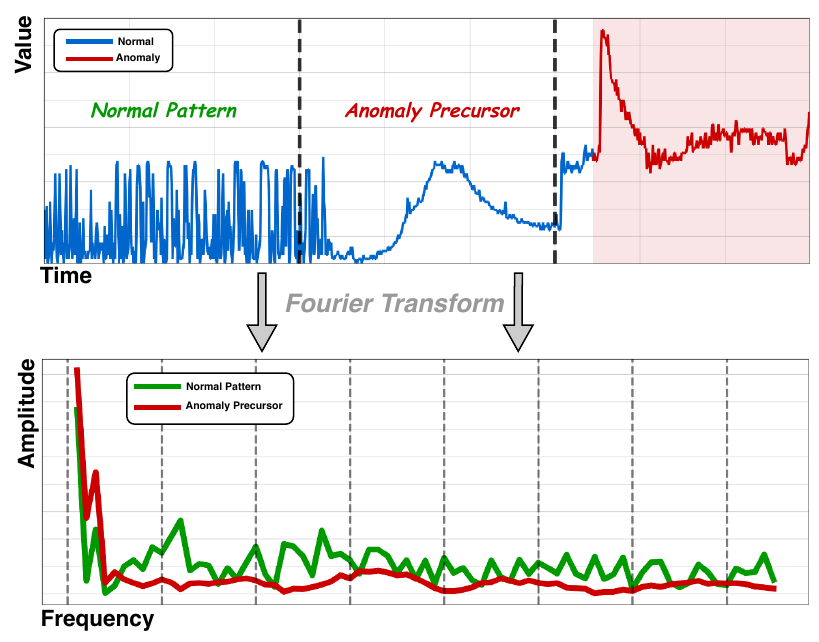}\label{fig:1a}}
    \hfill
    \subfigure[PSM]{\includegraphics[width=0.48\linewidth]{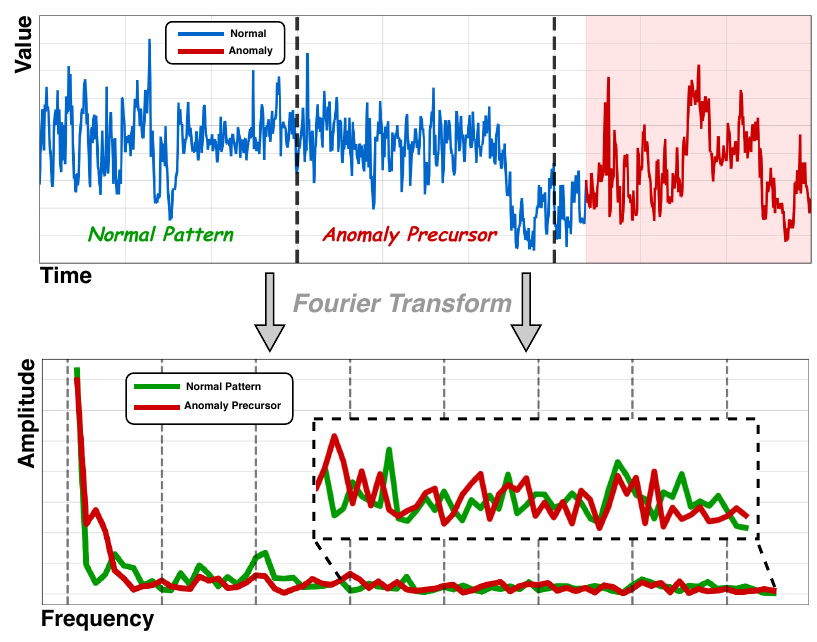}\label{fig:1b}}
    \caption{Anomaly precursors on SMD and PSM datasets. The red backgrounds indicate the actual anomaly segment. Prior to the occurrence of real anomalies, anomaly precursors have undergone a gradual change compared to the normal pattern and these changes are particularly apparent in the frequency domain.}
    \label{fig:1}  
\end{figure*}
 
In recent years, various unsupervised AP methods have been proposed\cite{zhao2024fcm,hu2024multirc,park2025a2p,jhin2023pad}, benefiting from the rapid development of deep learning. Previous studies have indicated that anomalies in production often do not occur suddenly; instead, they evolve incrementally, except in the case of deliberate attacks\cite{jhin2023pad,hu2024multirc}. Prior to actual anomalies, gradual pattern changes, defined as anomaly precursors, can be observed (Figure \ref{fig:1}).
Previous methods such as MultiRC\cite{hu2024multirc} and PAD\cite{jhin2023pad} predict future anomalies by identifying these anomaly precursors. Nonetheless, these approaches primarily focus on analyzing the underlying patterns of precursors in the time domain, overlooking valuable frequency-domain information. As shown in Figure \ref{fig:1}, anomaly precursors and normal patterns exhibit significant differences in the frequency domain, attributed to the gradual pattern changes preceding actual anomalies. Relying exclusively on time domain features may overlook the gradual changes of patterns, failing to identify anomaly precursors effectively.

Nevertheless, anomaly precursor identification methods are inadequate for real-world scenarios. MultiRC\cite{hu2024multirc} and PAD\cite{jhin2023pad} can only detect anomalies that may occur in the near term, while lacking the ability to output exact anomaly timestamps \cite{park2025a2p}. In the real world, it is crucial to predict anomalies across variable time horizons in which we are interested. Several recent approaches, such as FCM\cite{zhao2024fcm} and A2P\cite{park2025a2p}, predict anomalies through future context modeling, which involves predicting future data and subsequently detecting anomalies. Although these methods highlight the importance of time series forecasting, their forecasting models are trained on the traditional objective of next-series prediction, failing to support the prediction of strong anomalies from subtle precursors. Therefore, our intuition is that an effective training objective should be designed to guide the model to consider the influence of extended contextual information, strengthening its sensitivity to subtle anomaly precursors.

Furthermore, existing models are trained exclusively on normal data. When precursors exhibit extremely subtle deviations from normal data, forecasting models still reduce the degree of abnormality of anomaly time points, even when the new training objective is designed to enhance its sensitivity to such precursors. To address this limitation, it is imperative to amplify the attenuated anomalous signals in forecasts. This motivates us to establish a normal forecast baseline and  compare it with the current outputs, thus enhancing the predicted anomalous signal.

Motivated by the aforementioned observations, we propose the \textbf{R}econstruction-\textbf{E}limination-based \textbf{D}ual-stream Contrastive \textbf{F}orecasting (\textbf{RED-F}) framework, which is composed of two core components: the \textbf{R}econstruction-\textbf{E}limination \textbf{M}odel (\textbf{REM}) and the \textbf{D}ual-stream Contrastive \textbf{F}orecasting \textbf{M}odel (\textbf{DFM}). REM aims to consider the existence of normal patterns in the training process. When anomaly precursors are detected, such precursor signals can be captured and reconstructed to normal pattern, providing a foundation for subsequent contrastive forecasting. To achieve the object, we utilize Fourier Transformation\cite{brigham2021fft} to stretch across time and frequency domains to facilitate the detection of gradual pattern changes. To comprehensively capture normal patterns in the frequency domain, we adopt the frequency patching technique to generate diverse frequency bands and propose a dual-perspective modeling approach which considers both the intra-band and the inter-band relationships. Specifically, in modeling inter-band relationships, we constructed a relationship graph based on weighted differences of frequency-domain features to represent the frequency-band relationships between channels. Through comprehensively modeling  normal patterns in the frequency domain, we tackle the problem of detecting anomaly precursors that involve underlying pattern changes. Leveraging REM’s capability to detect underlying anomaly precursors, we propose DFM incorporated with Multi-Series Prediction (MSP) to improve the performance of anomaly prediction. Inspired by Multi-Token Prediction (MTP) \cite{liu2024deepseek}, MSP enhances the sensitivity of the forecasting models to anomaly precursors by fully integrating future context. We utilize MSP modules to generate multiple auxiliary predictions for distant future segments, sharing the embedding and output layers of the forecasting model. By co-optimizing all prediction tasks, the shared layers are driven to learn the influence of future context. Furthermore, we employ an MSP-trained backbone, enabling simultaneous forecast on both the original window and the reconstructed window. Finally, the future anomaly score can be obtained by computing the difference between the two forecast results.

Our main contributions are summarized as follows:
\begin{itemize}
    \item We propose RED-F, a novel AP framework, which transforms the challenging task of weak precursor detection into a simpler relative comparison. Extensive experiments demonstrate that it significantly outperforms twelve baseline models on six real-world datasets.
    \item We validate REM as a standalone reconstruction-based detector for traditional AD tasks. Extensive experiments show that REM outperforms existing state-of-the-art (SOTA) AD models.
    \item We demonstrate the effectiveness of DFM and its MSP-trained objective. When adapted for long-term time series forecasting (TSF), the DFM outperforms current SOTA TSF models.
\end{itemize}

\section{Related Work}
This section summarizes the related work on time series anomaly prediction, which can be primarily divided into two categories: methods for identifying anomaly precursors and methods for future context modeling.
\subsection{Methods for identifying anomaly precursors}

PAD\cite{jhin2023pad} first proposes Precursor-of-Anomaly (PoA) detection, which uses dual-stream neural controlled differential equations and knowledge distillation for simultaneous anomaly and precursor detection. For unsupervised training, PAD generates artificial anomalies by resampling normal data. However, this simple augmentation strategy fails to simulate the diversity of real-world precursors, limiting PAD's ability to handle varied precursor patterns.

To address the diversity, MultiRC\cite{hu2024multirc} introduces a multi-scale structure and an adaptive periodic mask, which allows the model to adjust its attended time scale based on dominant frequency domain periods, adapting to precursors of varying durations. It also injects diverse noise types to generate negative samples for contrastive learning. However,  Relying exclusively on time domain features
may overlook the gradual pattern changes, failing to detect gradually evolving anomaly precursors.

Furthermore, these methods merely provide a binary judgment on whether an anomaly will emerge soon, offering no information about its specific timing. Hence, they cannot be applied to scenarios where identifying time steps with anomalies is crucial.

\subsection{Methods for future context modeling}

These methods aim to predict anomalies by forecasting future values. FCM\cite{zhao2024fcm} adopts a scheme where it forecasts a target window from the historical window and concatenates them, evaluating the reconstruction error of the combined window. The key insight is that the reconstruction error of the combined window is expected to be large if the future target window contains abnormal time points, and it would be small otherwise. However, Forecasting directly from subtle precursors often yields a prediction that is overwhelmed by the normal pattern, and concatenating it with historical data introduces further noise.

To solve this problem of predictions being overwhelmed by normal patterns, \cite{park2025a2p} propose the A2P framework. They explicitly note that predictions from a model trained on normal data are easily submerged within the normal pattern. Therefore, A2P introduces the AAF and SAP modules to inject synthetic anomaly patterns into the training data, teaching the model to capture the evolutionary laws of anomalies and the relationship between precursors and future anomalies.

Despite injecting synthetic anomalies, the forecasting model of A2P is trained on traditional objective of next-series prediction. Because of the inherent weakness of precursors, their impact on the prediction is extremely limited. More importantly, single-stream prediction is still easily overwhelmed by the normal pattern, making it difficult to effectively amplify future anomaly signals.

\section{Preliminaries}
This section first introduces the concept of multivariate time series anomaly prediction, and then summarizes the definition of Fast Fourier Transform and Multi-Token Prediction.
\subsection{Multivariate Time Series Anomaly Prediction}
Time series anomaly prediction is a scenario designed to precisely pinpoint the exact time steps of anomaly points in the forthcoming signals.
Given an input signal $\mathbf{X_0} \in \mathbb{R}^{C \times L}$. The final goal is to obtain the binary results of future signals $\mathbf{S_0} \in \mathbb{R}^H$, where $L$ and $H$ are the lengths of the input and predicted signals, respectively, and $C$ is the number of channels in the signal.
\subsection{Fast Fourier Transform}
The Fast Fourier Transform (FFT)\cite{brigham2021fft} is an algorithm for efficiently computing the Discrete Fourier Transform (DFT) and its inverse. FFT transforms a signal from the time domain to the frequency domain. Given a discrete time series $X = \{x_0, x_1, \ldots, x_{L-1}\}$ of length $L$, its DFT is a complex sequence:
\begin{equation}
\begin{aligned}
\label{eq::1}
X_k = \sum_{n=0}^{L-1} x_n \cdot e^{-i \frac{2\pi kn}{L}}, \quad k = 0, 1, \ldots, L-1,
\end{aligned}
\end{equation}
where $X_k$ (for $k = 0, 1, \ldots, L-1$) is the complex component at frequency $\frac{k}{L}$, and its amplitude $|X_k|$ reflects its strength in the original signal. Similarly, the Inverse Discrete Fourier Transform (IDFT),  efficiently computed via the Inverse Fast Fourier Transform (iFFT), reconstructs the original time domain signal $x_n$ from its frequency representation $X_k$:
\begin{equation}
\label{eq::2}
x_n = \frac{1}{L} \sum_{k=0}^{L-1} X_k \cdot e^{i \frac{2\pi kn}{L}}, \quad n = 0, 1, \ldots, L-1.
\end{equation}

\begin{figure*}[!htbp]
    \centering
    \includegraphics[width=1.0\linewidth]{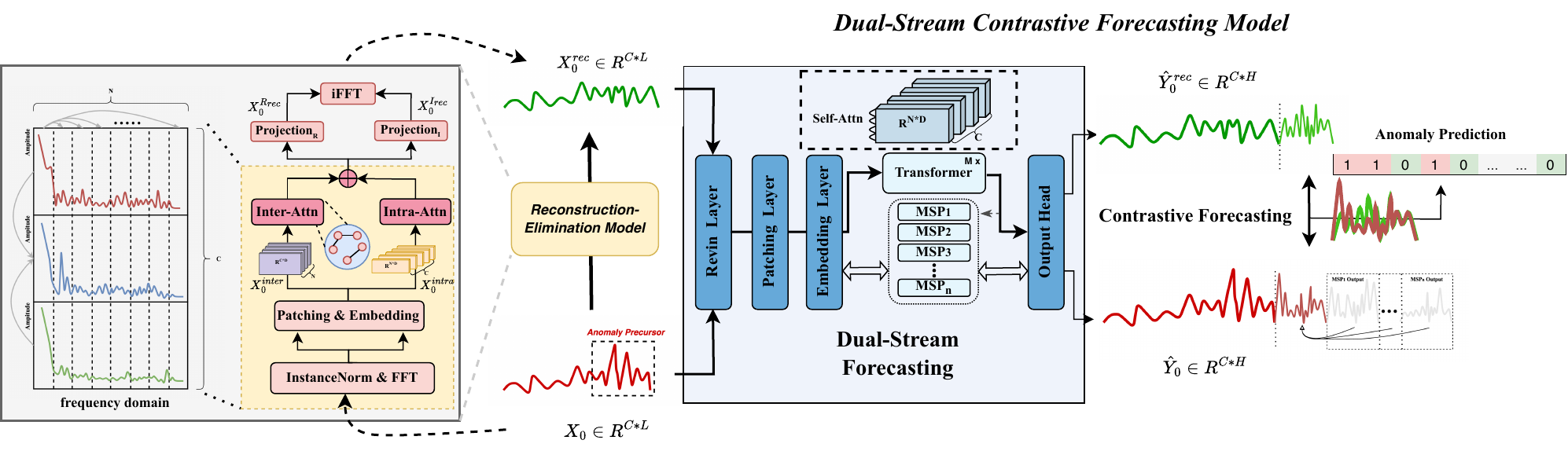}
    \caption{The architecture of RED-F. It consists of two components: (1) Reconstruction-Elimination Model (REM), which eliminates the subtle fluctuations in the anomaly precursor to obtain its normal pattern; (2) Dual-Stream Contrastive Forecasting Model (DFM), which predicts future anomalies based on the predictive comparison between the anomaly precursor and the normal pattern.}
    \label{fig:arc}
\end{figure*}
\subsection{Multi-Token Prediction}
MTP \cite{liu2024deepseek} extends next-token prediction by introducing $D$ auxiliary prediction depths. On the one hand, the MTP objective densifies training signals and can improve data efficiency. On the other hand, it encourages the model to pre-plan its representations, leading to better prediction of future tokens.

Concretely, the MTP implementation uses $D$ prediction modules to generate $D$ additional future tokens. 
The $k$-th MTP module consists of a shared embedding layer $Emb(\cdot)$, a shared output head $OutHead(\cdot)$, a Transformer block $TRM_k(\cdot)$, and a projection matrix $M_k \in \mathbb{R}^{d \times 2d}$. 
At the $k$-th prediction depth, the model first combines the previous hidden state $\mathbf{h}_i^{k-1} \in \mathbb{R}^{d}$ and the embedding of the $(i+k)$-th token $Emb(t_{i+k}) \in \mathbb{R}^{d}$ with a linear projection:
\begin{equation}
    \mathbf{h}_i^{\prime k} = M_k [RMSNorm(\mathbf{h}_i^{k-1}) ; \operatorname{RMSNorm}(\operatorname{Emb}(t_{i+k}))],
\end{equation}
where $[\cdot ; \cdot]$ denotes concatenation. 
Especially, when $k=1$, $h_i^{k-1}$ refers to the representation given by the main model.
Note that each MTP module, shares its embedding layer with the main model. 
The combined $h_i^{\prime k}$ is then fed into the Transformer block at the $k$-th depth to produce the output representation at the current depth $\mathbf{h}_{i}^{k}$:
\begin{equation}
    \mathbf{h}_{1:T-k}^{k} = TRM_k(\mathbf{h}_{1:T-k}^{\prime k}),
\end{equation}
where $T$ represents the input sequence length and $_{i:j}$ denotes the slicing operation (inclusive of both the left and right boundaries). 
Finally, given $h_{i}^{k}$ as input, the shared output head computes the probability distribution for the $k$-th additional predicted token $P_{i+k+1}^{k} \in \mathbb{R}^{V}$, where $V$ is the vocabulary size:
\begin{equation}
    P_{i+k+1}^{k} = OutHead(\mathbf{h}_{i}^{k}).
\end{equation}

The output head $OutHead(\cdot)$ first projects the representation to logits and then applies a $Softmax(\cdot)$ function to obtain the prediction probabilities for the $k$-th additional token. 
Also, for each MTP module, its output head is shared with the main model. 

In this paper, we adapt the MTP concept to design our MSP, aiming to enhance the sensitivity of model to anomaly precursors.

\section{Methodology}
This section overviews the architecture of RED-F and then details each model in the framework.
\subsection{Overview}
The overall architecture of RED-F is shown in Figure \ref{fig:arc}, which is composed of the Reconstruction-Elimination Model (REM) and the Dual-Stream Contrastive Forecasting Model (DFM). 
Given the input MTS data $\mathbf{X_0} \in \mathbb{R}^{C \times L}$, we first apply REM to reconstruct the time series to the normal pattern $\mathbf{X_0^{rec}} \in \mathbb{R}^{C \times L}$,  providing a normal reference for subsequent contrastive forecasting. Since anomaly precursors manifest gradually, REM employs FFT to map the time series into the frequency domain, using frequency patching and a dual-perspective modeling approach which considers both the intra-band and the inter-band relationships to perform fine-grained frequency
domain modeling. Subsequently, DFM receives both the original input window and the reconstructed window from REM simultaneously. These windows are fed into a shared backbone whose accuracy is enhanced by our proposed Multi-Series Prediction (MSP) to generate respective forecasts. Finally, future anomaly scores are obtained by amplifying potential anomaly signals through comparative forecasting. The detailed designs of each model are presented in Sections \ref{sec:b} and \ref{sec:c}.

\subsection{Reconstruction-Elimination Model}\label{sec:b}
\subsubsection{InstanceNorm \& FFT}
Real-world series often exhibit non-stationarity\cite{liu2022non}, leading to distribution drift. Therefore, we adopt instance normalization to mitigate the distribution shift between training and test data. Then, we utilize the efficient FFT to map time series into orthogonal trigonometric signals in the frequency domain, retaining both real and imaginary parts through $X_0^R, X_0^I = \text{FFT}(X_0)$, where $X_0, X_0^R, X_0^I \in \mathbb{R}^{C \times L}$.

\subsubsection{Patching \& Embedding}
For fine-grained frequency-domain modeling, we create frequency bands via a patching operation from two perspectives, formalized on the real part:
\begin{equation}
\begin{aligned}
P_0^R &= \{P_0^{R_1}, P_0^{R_2}, \ldots, P_0^{R_N}\} = \text{Patching}_{\text{inter}}(X_0^R),\\
G_0^R &= \{G_0^{R_1}, G_0^{R_2}, \ldots, G_0^{R_C}\} = \text{Patching}_{\text{intra}}(X_0^R),
\end{aligned}
\end{equation}
where $P_0^R \in R^{N \times C \times p}, G_0^R \in \mathbb{R}^{C \times N \times p}$ denote the patch result of the real part from inter-band and intra-band views, respectively. The imaginary part is processed similarly to obtain $P_0^I$ and $G_0^I$. $N = \lfloor\frac{L-p}{s}\rfloor + 1$ is the total patche number, where $p$ is the patch size, and $s$ is the patch stride.
Next, we concatenate the real and imaginary parts of each view (e.g., $P_0^R$ with $P_0^I$, and $G_0^R$ with $G_0^I$). Then the combined representations are projected into the hidden space through an embedding layer. This yields the final inter-band feature representation and intra-band feature representation $X_0^{\text{inter}} \in \mathbb{R}^{N \times C \times D}$, $X_0^{\text{intra}} \in \mathbb{R}^{C \times N \times D}$, respectively, where $D$ is the hidden dimension.
\begin{figure}[t]
    \centering
    \includegraphics[width=1.0\linewidth]{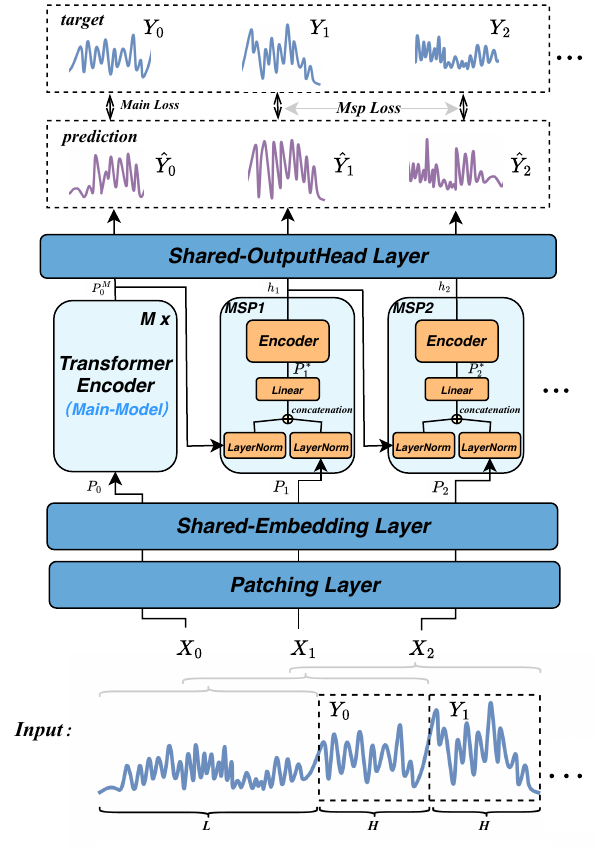}
    \caption{The architecture of DFM, illustrating how the main model is augmented by auxiliary MSP modules.}
    \label{fig:msp}
\end{figure}
\subsubsection{Inter-Attn}
To effectively model the complex dependencies of frequency-domain information among channels, we first generate a channel relationship graph by calculating the similarity of frequency-domain features. Subsequently, the graph is used to guide a standard self-attention module.
\paragraph{Relationship Graph Generation}
For the inter-band feature representation of the $i$-th frequency patch $X_0^{\text{inter}_i}$, we first compute its pointwise magnitude to obtain the magnitude matrix $F_i \in \mathbb{R}^{C \times D}$, where element $f_{i,jk}$ represents the magnitude of the $k$-th feature dimension of the $j$-th series in the $i$-th patch. Then we calculate the dissimilarity $d_{i,mn}$ between any two channels ($m, n$) using the weighted L1 distance between their frequency magnitude vectors. The dissimilarity is then inverted to produce a similarity $s_{i,mn}$, which can be formalized as :
\begin{equation}
\begin{aligned}
d_{i,mn} &= \sum_{k=1}^{D} w_k \cdot |f_{i,mk} - f_{i,nk}|,\\
s_{i,mn} &= \frac{1}{d_{i,mn} + \epsilon},
\end{aligned}
\end{equation}
where $m, n \in \{1, 2, \ldots, C\}$, $W = (w_1, \ldots, w_D) \in \mathbb{R}^D$ is a shared, learnable weight vector that adjusts the importance of different feature dimensions, and $\epsilon$ is a small constant for numerical stability. This process generates the similarity matrix $S_i \in \mathbb{R}^{C \times C}$ corresponding to $X_0^{\text{inter}_i}$. Finally, we perform differentiable Bernoulli sampling on $S_i$ to obtain $M_i$, which ensures that gradients can be backpropagated through the sampling process.

\paragraph{Masked Self-Attention}
After the inter-band relationship matrix $M_i$ is obtained, we utilize it to guide a standard Transformer encoder layer which includes a Multi-Head Self-Attention (MHSA) layer and a Feed-Forward Network (FFN), each with residual connections and layer normalization, used to further model the relationships between frequency bands. The computation for inter-band features of the $i$-th frequency patch $X_0^{\text{inter}_i}$ can be formalized as:
\begin{equation}
\begin{aligned}
&X_0^{\prime{\mathrm{inter}}_i} = \text{LayerNorm}(X_0^{\text{inter}_i}),\\
&O_0^{\text{inter}_i} = \text{MHSA}(X_0^{\prime{\mathrm{inter}}_i}, \text{mask}=M_i),\\
&X_0^{*{\mathrm{inter}}_i} = X_0^{\prime{\mathrm{inter}}_i} + O_0^{\text{inter}_i} ,\\
&\widetilde{X}_0^{\mathrm{inter}_i} = \text{FFN}(\text{LayerNorm}(X_0^{*{\mathrm{inter}}_i})) + X_0^{*{\mathrm{inter}}_i},\\
\end{aligned}
\end{equation}
where $M_i \in \mathbb{R}^{C \times C}$ is the similarity matrix used as a mask. The final output $\widetilde{X}_0^{\mathrm{inter}_i} \in \mathbb{R}^{C \times D}$ represents the processed features for the $i$-th patch. This process is repeated for all patches, yielding the updated inter-band feature representation $\widetilde{X}_0^{\text{inter}} \in \mathbb{R}^{N \times C \times D}$.

\begin{algorithm}[t]
\caption{Sample Generation for MSP-based Training}
\label{alg:msp_sample_generation}
\begin{algorithmic}[1]
    \STATE \textbf{Input:} Full time series $\mathbf{TS} \in \mathbb{R}^{C \times T_{\text{total}}}$, starting index $t_{\text{start}}$, look-back length $L$, horizon length $H$, MSP count $n$
    \STATE \textbf{Initialize:} $\mathcal{X} \gets \emptyset$, $\mathcal{Y} \gets \emptyset$
    \FOR{$k = 0$ \TO $n$} 
        \STATE $input\_start \gets t_{\text{start}} + k \cdot H$
        \STATE $input\_end \gets input\_start + L$
        \STATE $\mathbf{X}_k \gets \mathbf{TS}[:, input\_start : input\_end]$
        \STATE $target\_start \gets input\_end$
        \STATE $target\_end \gets target\_start + H$
        \STATE $\mathbf{Y}_k \gets \mathbf{TS}[:, target\_start : target\_end]$ 
        \STATE Append $\mathbf{X}_k$ to $\mathcal{X}$
        \STATE Append $\mathbf{Y}_k$ to $\mathcal{Y}$
    \ENDFOR
    \STATE \textbf{Output:} $\mathcal{X} = \{\mathbf{X}_0, \dots, \mathbf{X}_n\}$, $\mathcal{Y} = \{\mathbf{Y}_0, \dots, \mathbf{Y}_n\}$
\end{algorithmic}
\end{algorithm}

\subsubsection{Intra-Attn}
For the intra-series perspective, we process the feature representation $X_0^{\text{intra}_i}$ for the $i$-th channel. Because there is no temporal causality between the different frequency bands, we directly use a standard Transformer encoder layer to capture their relationships. The computation is similar to the second stage of Inter-Attn but without the masking operation, obtaining the updated intra-band representation $\widetilde{X}_0^{\text{intra}}$.
\subsubsection{Projection \& iFFT}
After modeling from both perspectives, the output features $\widetilde{X}_0^{\text{inter}}$ and $\widetilde{X}_0^{\text{intra}}$ are fused. We first permute the dimensions of $\widetilde{X}_0^{\text{inter}}$ to align its shape with $\widetilde{X}_0^{\text{intra}}$, then fuse them to obtain a unified feature representation $\widetilde{X}_0$. This unified representation $\widetilde{X}_0$ is then decoded back into the real and imaginary parts, achieved by flattening the features and passing them through two independent projection MLPs:
\begin{equation}
\begin{aligned}
&X_0^{R_{rec}} = \text{Projection}_R(\widetilde{X}_0) = \text{MLP}_R(\text{Flatten}(\widetilde{X}_0)),   \\
&X_0^{I_{rec}} = \text{Projection}_I(\widetilde{X}_0) = \text{MLP}_I(\text{Flatten}(\widetilde{X}_0)),  \\
\end{aligned}
\end{equation}
where $X_0^{R_{rec}}, X_0^{I_{rec}} \in \mathbb{R}^{C \times L}$ are the reconstructed real ($X^R$) and imaginary ($X^I$) parts, respectively. Finally, we obtain temporal reconstruction $X_0^{\text{rec}} \in \mathbb{R}^{C \times L}$ through iFFT.

\subsection{Dual-Stream Contrastive Forecasting Model}\label{sec:c}
\subsubsection{Main Model}
The input data $X_0$ is passed through a Revin layer, followed by Patching and shared Embedding layers to generate a sequence of patches $P_0 \in \mathbb{R}^{C \times N \times D}$ defined in Eq \ref{forecasting_patching}, where $N$ is the number of patches and $D$ is the hidden dimension. Then the patches are processed by a deep neural network composed of $M$ stacked Encoder layers, each with a residual connection. We take the $m$-th Encoder layer as an example:
\begin{equation}
\label{forecasting_patching}
P_0 = \text{Embedding}(\text{Patching}(\text{Revin}(X_0))),
\end{equation}
\begin{equation}
P_0^m = P_0^{m-1} + \text{Encoder}^m(P_0^{m-1}).
\end{equation}

Afterwards it generates the representation denoted as $P_0^M \in \mathbb{R}^{C \times N \times D}$. Finally a share OutputHead layer is used to obtain the prediction result $\hat{Y}_0 \in \mathbb{R}^{C \times H}$.
Additionally, the reconstructed input data $X_0^{\text{rec}}$ is passed through the main model simultaneously to obtain its corresponding prediction result $\hat{Y}_0^{\text{rec}}$.

\begin{table*}[htbp]
\centering
\caption{Statistics of multivariate datasets selected for Anomaly Prediction (AP) evaluation (AR: anomaly ratio).}
\label{tab:data}
\begin{small}
\setlength{\tabcolsep}{5pt}
\begin{tabularx}{\textwidth}{@{}l l r r r r X@{}}
\toprule
\textbf{Dataset} & \textbf{Domain} & \textbf{Variable} &
\textbf{Total} &\textbf{Test} &
\textbf{AR~(\%)} &
\textbf{Description} \\
\midrule
MSL     & Spacecraft     & 55  & 132{,}046  & 73{,}729  & 5.88  & Spacecraft telemetry and anomaly data from the MSL rover. \\
PSM     & Server Machine & 25  & 220{,}322  & 87{,}841  & 11.07 & Application server telemetry data with labeled anomalies. \\
SMD     & Server Machine & 38  & 1{,}416{,}825 & 708{,}420 & 2.08  & Multivariate telemetry from 28 server machines. \\
GECCO   & Water Treatment & 9  & 138{,}521  & 69{,}261  & 1.25  & Water treatment data from the GECCO Challenge. \\
CICIDS  & Web Traffic    & 72  & 170{,}231  & 85{,}116  & 1.28  & Network traffic data with 80+ features and attack labels. \\
Genesis & Machinery      & 18  & 16{,}220   & 12{,}616  & 0.31  & Cyber-physical machinery sensor data. \\
\bottomrule
\end{tabularx}
\end{small}
\end{table*}

\subsubsection{Multi-Series Prediction}
Figure \ref{fig:msp} shows the detailed architecture of DFM, which is augmented by auxiliary Multi-Series Prediction (MSP) modules. To support our designed MSP-based training objective, i.e., to equip the model with a longer-term prediction horizon, we need a specific data generation method during training. As detailed in Algorithm \ref{alg:msp_sample_generation}, we generate a set of $n+1$ input-target pairs from a continuous time series, where $n$ represents the number of MSP modules during the training. Specifically, $X_0$ and $X_k$ are the input data for the main model and $k$-th MSP module, respectively. $Y_k$ is the target window corresponding to the input $X_k$. Notably, MSP auxiliary prediction is conducted exclusively on the original input data, excluding the reconstructed data. This will reduce the model's sensitivity to normal patterns, leading to a more stable prediction for the reconstructed window. The $n$ MSP modules are used to predict the more distant future ($Y_1, \ldots, Y_n$). For the $k$-th MSP module, its input $X_k$ first passes through the shared embedding layer to obtain $P_k$. Next the $k$-th MSP module receives the representation from the preceding module (either the main model or the $(k-1)$-th MSP module). Both are normalized via a LayerNorm layer, then concatenated, and finally combined using a Linear layer:
\begin{align}
{\widetilde{P}}_k = \text{LayerNorm}(P_k), \hspace{1em}& 
{\widetilde{h}}_{k-1} = \text{LayerNorm}(h_{k-1}),
\end{align}
\begin{equation}
P_k^* = \text{Linear}(\text{Concatenation}({\widetilde{P}}_k, {\widetilde{h}}_{k-1}, \text{dim}=-1)),
\end{equation}
where $P_k \in \mathbb{R}^{C \times N \times D}$ is the patch embedding result of the $k$-th input window, and $h_{k-1} \in \mathbb{R}^{C \times N \times D}$ is the representation of the previous module. Especially, when $k$=1, $h_{k-1}$ refers to the representation $P_0^M$ given by the main model. $P_k^* \in \mathbb{R}^{C \times N \times D}$ is the combined representation. 

Finally, $P_k^*$ is passed through only one Encoder layer and the shared OutputHead layer is used to obtain the $k$-th prediction result $\hat{Y}_k \in \mathbb{R}^{C \times H}$:
\begin{align}
h_k = \text{Encoder}(P_k^*), \hspace{1em}& 
\hat{Y}_k = \text{OutputHead}(h_k).
\end{align}

\begin{algorithm}[t]
\caption{RED-F Joint Training}
\label{alg:redf_training}
\begin{algorithmic}[1]
    \STATE \textbf{Input:} Training dataset $\mathcal{D}$, batch size $B$, MSP count $n$, Look-back $L$, Horizon $H$
    \STATE \textbf{Initialize:} REM parameters $\theta_{rem}$, DFM parameters $\theta_{dfm}$
    \STATE \textbf{Initialize:} Loss weights $\{\lambda_{\text{time}}, \lambda_{\text{freq}}, \lambda_{\text{main}}, \lambda_{\text{msp}}, \lambda_{\text{contra}}\}$

    \WHILE{not converged}
        \STATE \textit{// Batch Sample Generation (Algorithm 1)}
        \STATE Sample batch of starting indices $\{t_{\text{start}}^{(i)}\}_{i=1}^{B}$ from $\mathcal{D}$
        \STATE Generate MSP samples $\{(\mathbf{X}_k^{(i)}, \mathbf{Y}_k^{(i)})\}_{k=0}^{n}$ for each $t_{\text{start}}^{(i)}$
        \STATE
        
        \STATE \textit{// Reconstruction-Elimination Model (REM)}
        \STATE $\mathbf{X}_{0}^{rec}, \mathbf{X}_{rec}^{R}, \mathbf{X}_{rec}^{I} \gets \text{REM}(\mathbf{X}_{0}; \theta_{rem})$
        \STATE \textit{// REM Loss Calculation [Eq. 16-18]}
        \STATE $\mathcal{L}_{rem} = \lambda_{time} \cdot \mathcal{L}_{time} + \lambda_{freq} \cdot \mathcal{L}_{freq}$
        \STATE $\mathcal{L}_{time} = ||X_0 - X^{rec}_0||^2$
        \STATE $\mathcal{L}_{freq} = ||X^R_0 - X_0^{R_{rec}}|| + ||X^I_0 - X_0^{I_{rec}}||$
        \STATE

        \STATE \textit{// Dual-Stream Contrastive Forecasting (DFM)}
        \STATE $\hat{\mathbf{Y}}_{0} \gets \text{DFM}_{\text{main}}(\mathbf{X}_{0}; \theta_{dfm})$
        \STATE $\hat{\mathbf{Y}}_{0}^{rec} \gets \text{DFM}_{\text{main}}(\mathbf{X}_{0}^{rec}; \theta_{dfm})$ 
        \STATE $\{\hat{\mathbf{Y}}_{1}, \dots, \hat{\mathbf{Y}}_{n}\} \gets \text{DFM}_{\text{msp}}(\{\mathbf{X}_{1}, \dots, \mathbf{X}_{n}\}; \theta_{dfm})$ 
        \STATE \textit{// DFM Loss Calculation [Eq. 19-22]}
        \STATE $\mathcal{L}_{dfm} = \lambda_{main} \cdot \mathcal{L}_{main}+\lambda_{msp} \cdot \mathcal{L}_{msp} + \lambda_{contra} \cdot \mathcal{L}_{contra}$
        \STATE $\mathcal{L}_{main}=||Y_0 - \hat{Y}_0||^2$
        \STATE $\mathcal{L}_{msp}=\sum_{k=1}^{n} ||Y_k - \hat{Y}_k||^2$
        \STATE $\mathcal{L}_{contra}=||\hat{Y}_0 - \hat{Y}^{rec}_0||^2$
        \STATE

        \STATE \textit{// Joint Optimization  \textit{[Eq. 15]}}
        \STATE $\mathcal{L} \gets \mathcal{L}_{rem} + \mathcal{L}_{dfm}$
        \STATE Compute gradients $\nabla_{\theta_{rem}, \theta_{dfm}} \mathcal{L}$
        \STATE Update parameters $(\theta_{rem}, \theta_{dfm})$ using optimizer
    \ENDWHILE

    \STATE \textbf{Output:} Optimized parameters $\theta_{rem}, \theta_{dfm}$
\end{algorithmic}
\end{algorithm}

\subsection{Total Objective Function}
The total objective function of our proposed framework is summarized as follows:
\begin{equation}
\mathcal{L} = \mathcal{L}_{rem} + \mathcal{L}_{dfm}. 
\end{equation}

\subsubsection{REM Objective Function}
$\mathcal{L}_{rem}$ is composed of the reconstruction functions in time and frequency domains:
\begin{equation}
\mathcal{L}_{rem} = \lambda_{time} \cdot \mathcal{L}_{time} + \lambda_{freq} \cdot \mathcal{L}_{freq},
\end{equation}
\begin{equation}
\mathcal{L}_{time} = ||X_0 - X^{rec}_0||^2,
\end{equation}
\begin{equation}
\mathcal{L}_{freq} = ||X^R_0 - X_0^{R_{rec}}|| + ||X^I_0 - X_0^{I_{rec}}||.
\end{equation}

We utilize 2-norm and 1-norm to distinguish the distinct numerical characteristics of time and frequency domains.
\subsubsection{DFM Objective Function}
$\mathcal{L}_{dfm}$ is composed of the functions of forecasting with MSP and contrastive forecasting:
\begin{equation}
\mathcal{L}_{dfm} = \lambda_{main} \cdot \mathcal{L}_{main}+\lambda_{msp} \cdot \mathcal{L}_{msp} + \lambda_{contra} \cdot \mathcal{L}_{contra},
\end{equation}
\begin{equation}
\mathcal{L}_{main}=||Y_0 - \hat{Y}_0||^2,
\end{equation}
\begin{equation}
\mathcal{L}_{msp}=\sum_{k=1}^{n} ||Y_k - \hat{Y}_k||^2,
\end{equation}
\begin{equation}
\mathcal{L}_{contra}=||\hat{Y}_0 - \hat{Y}^{rec}_0||^2.
\end{equation}

Algorithm \ref{alg:redf_training} provides a comprehensive overview of how the REM and DFM are jointly optimized in an end-to-end manner.
\begin{table*}[ht]
  \vspace{-5pt}
  \renewcommand{\arraystretch}{1.25} 
  \centering
  \begin{threeparttable}
  \begin{small}
  \renewcommand{\multirowsetup}{\centering}
  \setlength{\tabcolsep}{1.55pt}
  \caption{Anomaly prediction results with input length $L=192$ and prediction horizons $H = \{32, 64, 96, 128\}$. Best results are in \textbf{bold}.}
  \label{tab:main-result}
  \begin{tabular}{c|c|cccc|cccc|cccc|cccc|cccc|ccccc}
    \toprule
    \multicolumn{2}{c}{\rotatebox{0}{\scalebox{1.05}Dataset}} &
    \multicolumn{4}{c}{\rotatebox{0}{\scalebox{1}{MSL}}} &
    \multicolumn{4}{c}{\rotatebox{0}{\scalebox{1}{PSM}}} &

    \multicolumn{4}{c}{\rotatebox{0}{\scalebox{1}{SMD}}} &
    \multicolumn{4}{c}{\rotatebox{0}{\scalebox{1}{GECCO}}} &
    \multicolumn{4}{c}{\rotatebox{0}{\scalebox{1}{CICIDS}}} &
    \multicolumn{4}{c}{\rotatebox{0}{\scalebox{1}{Genesis}}}
 \\
    \cmidrule(lr){3-6}\cmidrule(lr){7-10}\cmidrule(lr){11-14}\cmidrule(lr){15-18}\cmidrule(lr){19-22} \cmidrule(lr){23-26}
    \multicolumn{2}{c}{\scalebox{1}{H}}  & \scalebox{0.78}{32} & \scalebox{0.78}{64}  & \scalebox{0.78}{96} & \scalebox{0.78}{128}  & \scalebox{0.78}{32} & \scalebox{0.78}{64}  & \scalebox{0.78}{96} & \scalebox{0.78}{128}& \scalebox{0.78}{32} & \scalebox{0.78}{64}  & \scalebox{0.78}{96} & \scalebox{0.78}{128}& \scalebox{0.78}{32} & \scalebox{0.78}{64}  & \scalebox{0.78}{96} & \scalebox{0.78}{128}& \scalebox{0.78}{32} & \scalebox{0.78}{64}  & \scalebox{0.78}{96} & \scalebox{0.78}{128}& \scalebox{0.78}{32} & \scalebox{0.78}{64}  & \scalebox{0.78}{96} & \scalebox{0.78}{128}\\
    \toprule
    \multicolumn{2}{c|}{\scalebox{0.9}{DLinear+AT}}&\scalebox{0.78}{0.663}&\scalebox{0.78}{0.654}&\scalebox{0.78}{0.657}&\scalebox{0.78}{0.643}&\scalebox{0.78}{0.600}&\scalebox{0.78}{0.654}&\scalebox{0.78}{0.655}&\scalebox{0.78}{0.653}&\scalebox{0.78}{0.667}&\scalebox{0.78}{0.659}&\scalebox{0.78}{0.662}&\scalebox{0.78}{0.657}&\scalebox{0.78}{0.788}&\scalebox{0.78}{0.779}&\scalebox{0.78}{0.759}&\scalebox{0.78}{0.662}&\scalebox{0.78}{0.527}&\scalebox{0.78}{0.441}&\scalebox{0.78}{0.560}&\scalebox{0.78}{0.378}&\scalebox{0.78}{0.679}&\scalebox{0.78}{0.689}&\scalebox{0.78}{0.734}&\scalebox{0.78}{0.665}\\
    \multicolumn{2}{c|}{\scalebox{0.9}{DLinear+DC}}
     &\scalebox{0.78}{0.659}
     &\scalebox{0.78}{0.661}
     &\scalebox{0.78}{0.667}
     &\scalebox{0.78}{0.664}
     &\scalebox{0.78}{0.628}
     &\scalebox{0.78}{0.620}
     &\scalebox{0.78}{0.610}
     &\scalebox{0.78}{0.559}
     &\scalebox{0.78}{0.602}
     &\scalebox{0.78}{0.610}
     &\scalebox{0.78}{0.546}
     &\scalebox{0.78}{0.578}
     &\scalebox{0.78}{0.751}
     &\scalebox{0.78}{0.782}
     &\scalebox{0.78}{0.739}
     &\scalebox{0.78}{0.744}
     &\scalebox{0.78}{0.578}
     &\scalebox{0.78}{0.556}
     &\scalebox{0.78}{0.594}
     &\scalebox{0.78}{0.466}
     &\scalebox{0.78}{0.710}
     &\scalebox{0.78}{0.708}
     &\scalebox{0.78}{0.743}
     &\scalebox{0.78}{0.703}
    \\
    \multicolumn{2}{c|}{\scalebox{0.9}{Nlinear+AT}}
     &\scalebox{0.78}{0.658}
     &\scalebox{0.78}{0.645}
     &\scalebox{0.78}{0.650}
     &\scalebox{0.78}{0.670}
     &\scalebox{0.78}{0.684}
     &\scalebox{0.78}{0.653}
     &\scalebox{0.78}{0.641}
     &\scalebox{0.78}{0.658}
     &\scalebox{0.78}{0.669}
     &\scalebox{0.78}{0.660}
     &\scalebox{0.78}{0.655}
     &\scalebox{0.78}{0.670}
     &\scalebox{0.78}{0.658}
     &\scalebox{0.78}{0.746}
     &\scalebox{0.78}{0.728}
     &\scalebox{0.78}{0.740}
     &\scalebox{0.78}{0.481}
     &\scalebox{0.78}{0.474}
     &\scalebox{0.78}{0.428}
     &\scalebox{0.78}{0.427}
     &\scalebox{0.78}{0.797}
     &\scalebox{0.78}{0.784}
     &\scalebox{0.78}{0.736}
     &\scalebox{0.78}{0.582}
    \\
    \multicolumn{2}{c|}{\scalebox{0.9}{Nlinear+DC}}
     &\scalebox{0.78}{0.660}
     &\scalebox{0.78}{0.640}
     &\scalebox{0.78}{0.675}
     &\scalebox{0.78}{0.662}
     &\scalebox{0.78}{0.648}
     &\scalebox{0.78}{0.672}
     &\scalebox{0.78}{0.625}
     &\scalebox{0.78}{0.619}
     &\scalebox{0.78}{0.606}
     &\scalebox{0.78}{0.607}
     &\scalebox{0.78}{0.606}
     &\scalebox{0.78}{0.588}
     &\scalebox{0.78}{0.659}
     &\scalebox{0.78}{0.665}
     &\scalebox{0.78}{0.739}
     &\scalebox{0.78}{0.689}
     &\scalebox{0.78}{0.616}
     &\scalebox{0.78}{0.609}
     &\scalebox{0.78}{0.578}
     &\scalebox{0.78}{0.579}
     &\scalebox{0.78}{0.683}
     &\scalebox{0.78}{0.686}
     &\scalebox{0.78}{0.752}
     &\scalebox{0.78}{0.558}
    \\
    \multicolumn{2}{c|}{\scalebox{0.9}{PatchTST+AT}}
     &\scalebox{0.78}{0.653}
     &\scalebox{0.78}{0.661}
     &\scalebox{0.78}{0.652}
     &\scalebox{0.78}{0.658}
     &\scalebox{0.78}{0.681}
     &\scalebox{0.78}{0.652}
     &\scalebox{0.78}{0.679}
     &\scalebox{0.78}{0.636}
     &\scalebox{0.78}{0.663}
     &\scalebox{0.78}{0.662}
     &\scalebox{0.78}{0.671}
     &\scalebox{0.78}{0.660}
     &\scalebox{0.78}{0.566}
     &\scalebox{0.78}{0.751}
     &\scalebox{0.78}{0.752}
     &\scalebox{0.78}{0.732}
     &\scalebox{0.78}{0.517}
     &\scalebox{0.78}{0.437}
     &\scalebox{0.78}{0.455}
     &\scalebox{0.78}{0.501}
     &\scalebox{0.78}{0.776}
     &\scalebox{0.78}{0.786}
     &\scalebox{0.78}{0.739}
     &\scalebox{0.78}{0.662}
    \\
    
    \multicolumn{2}{c|}{\scalebox{0.9}{PatchTST+DC}}
     &\scalebox{0.78}{0.654}
     &\scalebox{0.78}{0.649}
     &\scalebox{0.78}{0.660}
     &\scalebox{0.78}{0.663}
     &\scalebox{0.78}{0.645}
     &\scalebox{0.78}{0.644}
     &\scalebox{0.78}{0.640}
     &\scalebox{0.78}{0.658}
     &\scalebox{0.78}{0.650}
     &\scalebox{0.78}{0.632}
     &\scalebox{0.78}{0.634}
     &\scalebox{0.78}{0.607}
     &\scalebox{0.78}{0.681}
     &\scalebox{0.78}{0.701}
     &\scalebox{0.78}{0.676}
     &\scalebox{0.78}{0.693}
     &\scalebox{0.78}{0.590}
     &\scalebox{0.78}{0.566}
     &\scalebox{0.78}{0.638}
     &\scalebox{0.78}{0.619}
     &\scalebox{0.78}{0.699}
     &\scalebox{0.78}{0.612}
     &\scalebox{0.78}{0.795}
     &\scalebox{0.78}{0.647}
     \\
    
     \multicolumn{2}{c|}{\scalebox{0.9}{TimesNet+AT}}
     &\scalebox{0.78}{0.651}
     &\scalebox{0.78}{0.665}
     &\scalebox{0.78}{0.683}
     &\scalebox{0.78}{0.672}
     &\scalebox{0.78}{0.656}
     &\scalebox{0.78}{0.652}
     &\scalebox{0.78}{0.625}
     &\scalebox{0.78}{0.626}
     &\scalebox{0.78}{0.661}
     &\scalebox{0.78}{0.661}
     &\scalebox{0.78}{0.670}
     &\scalebox{0.78}{0.666}
     &\scalebox{0.78}{0.783}
     &\scalebox{0.78}{0.781}
     &\scalebox{0.78}{0.755}
     &\scalebox{0.78}{0.692}
     &\scalebox{0.78}{0.566}
     &\scalebox{0.78}{0.491}
     &\scalebox{0.78}{0.602}
     &\scalebox{0.78}{0.504}
     &\scalebox{0.78}{0.786}
     &\scalebox{0.78}{0.801}
     &\scalebox{0.78}{0.697}
     &\scalebox{0.78}{0.603}
    \\

    \multicolumn{2}{c|}{\scalebox{0.9}{TimesNet+DC}}
     &\scalebox{0.78}{0.669}
     &\scalebox{0.78}{0.686}
     &\scalebox{0.78}{0.668}
     &\scalebox{0.78}{0.671}
     &\scalebox{0.78}{0.629}
     &\scalebox{0.78}{0.651}
     &\scalebox{0.78}{0.640}
     &\scalebox{0.78}{0.618}
     &\scalebox{0.78}{0.657}
     &\scalebox{0.78}{0.659}
     &\scalebox{0.78}{0.666}
     &\scalebox{0.78}{0.669}
     &\scalebox{0.78}{0.728}
     &\scalebox{0.78}{0.761}
     &\scalebox{0.78}{0.758}
     &\scalebox{0.78}{0.730}
     &\scalebox{0.78}{0.600}
     &\scalebox{0.78}{0.556}
     &\scalebox{0.78}{0.583}
     &\scalebox{0.78}{0.556}
     &\scalebox{0.78}{0.772}
     &\scalebox{0.78}{0.799}
     &\scalebox{0.78}{0.745}
     &\scalebox{0.78}{0.658}
    \\

    \multicolumn{2}{c|}{\scalebox{0.9}{iTransformer+AT}}
     &\scalebox{0.78}{0.654}
     &\scalebox{0.78}{0.663}
     &\scalebox{0.78}{0.671}
     &\scalebox{0.78}{0.679}
     &\scalebox{0.78}{0.657}
     &\scalebox{0.78}{0.630}
     &\scalebox{0.78}{0.639}
     &\scalebox{0.78}{0.650}
     &\scalebox{0.78}{0.659}
     &\scalebox{0.78}{0.659}
     &\scalebox{0.78}{0.658}
     &\scalebox{0.78}{0.661}
     &\scalebox{0.78}{0.767}
     &\scalebox{0.78}{0.790}
     &\scalebox{0.78}{0.710}
     &\scalebox{0.78}{0.645}
     &\scalebox{0.78}{0.497}
     &\scalebox{0.78}{0.512}
     &\scalebox{0.78}{0.589}
     &\scalebox{0.78}{0.463}
     &\scalebox{0.78}{0.686}
     &\scalebox{0.78}{0.786}
     &\scalebox{0.78}{0.816}
     &\scalebox{0.78}{0.619}
     \\
     
    \multicolumn{2}{c|}{\scalebox{0.9}{iTransformer+DC}}
     &\scalebox{0.78}{0.681}
     &\scalebox{0.78}{0.676}
     &\scalebox{0.78}{0.675}
     &\scalebox{0.78}{0.676}
     &\scalebox{0.78}{0.683}
     &\scalebox{0.78}{0.540}
     &\scalebox{0.78}{0.578}
     &\scalebox{0.78}{0.541}
     &\scalebox{0.78}{0.653}
     &\scalebox{0.78}{0.630}
     &\scalebox{0.78}{0.660}
     &\scalebox{0.78}{0.649}
     &\scalebox{0.78}{0.692}
     &\scalebox{0.78}{0.579}
     &\scalebox{0.78}{0.658}
     &\scalebox{0.78}{0.634}
     &\scalebox{0.78}{0.617}
     &\scalebox{0.78}{0.587}
     &\scalebox{0.78}{0.550}
     &\scalebox{0.78}{0.603}
     &\scalebox{0.78}{0.794}
     &\boldres{\scalebox{0.78}{0.807}}
     &\scalebox{0.78}{0.697}
     &\scalebox{0.78}{0.681}
     \\

    \multicolumn{2}{c|}{\scalebox{0.9}{A2P}}
     &\scalebox{0.78}{0.690}
     &\scalebox{0.78}{0.630}
     &\scalebox{0.78}{0.640}
     &\scalebox{0.78}{0.619}
     &\scalebox{0.78}{0.636}
     &\scalebox{0.78}{0.679}
     &\scalebox{0.78}{0.632}
     &\scalebox{0.78}{0.623}
     &\scalebox{0.78}{0.582}
     &\scalebox{0.78}{0.574}
     &\scalebox{0.78}{0.563}
     &\scalebox{0.78}{0.537}
     &\scalebox{0.78}{0.660}
     &\scalebox{0.78}{0.626}
     &\scalebox{0.78}{0.612}
     &\scalebox{0.78}{0.587}
     &\scalebox{0.78}{0.567}
     &\scalebox{0.78}{0.536}
     &\scalebox{0.78}{0.522}
     &\scalebox{0.78}{0.500}
     &\scalebox{0.78}{0.720}
     &\scalebox{0.78}{0.700}
     &\scalebox{0.78}{0.687}
     &\scalebox{0.78}{0.677}
    \\
    \multicolumn{2}{c|}{\scalebox{0.9}{FCM}}
     &\scalebox{0.78}{0.673}
     &\scalebox{0.78}{0.684}
     &\scalebox{0.78}{0.681}
     &\boldres{\scalebox{0.78}{0.690}}
     &\scalebox{0.78}{0.706}
     &\scalebox{0.78}{0.693}
     &\scalebox{0.78}{0.694}
     &\scalebox{0.78}{0.705}
     &\scalebox{0.78}{0.676}
     &\scalebox{0.78}{0.679}
     &\scalebox{0.78}{0.681}
     &\scalebox{0.78}{0.685}
     &\scalebox{0.78}{0.676}
     &\scalebox{0.78}{0.682}
     &\scalebox{0.78}{0.710}
     &\scalebox{0.78}{0.739}
     &\scalebox{0.78}{0.565}
     &\scalebox{0.78}{0.580}
     &\scalebox{0.78}{0.537}
     &\scalebox{0.78}{0.495}
     &\scalebox{0.78}{0.731}
     &\scalebox{0.78}{0.694}
     &\scalebox{0.78}{0.659}
     &\scalebox{0.78}{0.660}
    \\
    
    \midrule
    \multicolumn{2}{c|}{\scalebox{1.0}{\textbf{\MakeUppercase{RED-F}}}}

     &\boldres{\scalebox{0.78}{0.698}}
     &\boldres{\scalebox{0.78}{0.695}}
     &\boldres{\scalebox{0.78}{0.684}}
     &\scalebox{0.78}{0.651}
     &\boldres{\scalebox{0.78}{0.759}}
     &\boldres{\scalebox{0.78}{0.756}}
     &\boldres{\scalebox{0.78}{0.717}}
     &\boldres{\scalebox{0.78}{0.735}}
     &\boldres{\scalebox{0.78}{0.735}}
     &\boldres{\scalebox{0.78}{0.723}}
     &\boldres{\scalebox{0.78}{0.724}}
     &\boldres{\scalebox{0.78}{0.702}}
     &\boldres{\scalebox{0.78}{0.791}}
     &\boldres{\scalebox{0.78}{0.799}}
     &\boldres{\scalebox{0.78}{0.780}}
     &\boldres{\scalebox{0.78}{0.750}}
     &\boldres{\scalebox{0.78}{0.641}}
     &\boldres{\scalebox{0.78}{0.649}}
     &\boldres{\scalebox{0.78}{0.644}}
     &\boldres{\scalebox{0.78}{0.637}}
     &\boldres{\scalebox{0.78}{0.821}}
     &\scalebox{0.78}{0.791}
     &\boldres{\scalebox{0.78}{0.823}}
     &\boldres{\scalebox{0.78}{0.847}}
     \\

    \bottomrule
  \end{tabular}
    \end{small}
  \end{threeparttable}
\vspace{-10pt}
\end{table*}

\subsection{Anomaly Criterion}
The future anomaly score is obtained using the contrastive forecasting result:
\begin{equation}
S_{\text{anomaly}} = \begin{bmatrix}||\hat{Y}_{0,:,i} -\hat{Y}_{0,:,i}^{\text{rec}}||^2\end{bmatrix}_{i=1,\ldots,H}
\end{equation}
where $S_{\text{anomaly}} \in \mathbb{R}^{H \times 1}$ denotes the point-wise anomaly criterion of the future window.

Following existing works\cite{wang2023d3r,shentu2024dada,li2025crossad}, after obtaining the anomaly score, we run SPOT\cite{siffer2017spot} to automatically compute the threshold $\delta$, and a point is marked as an anomaly if its anomaly score exceeds $\delta$.

\section{Experiments}

\subsection{Datasets} Unlike traditional Anomaly Detection (AD), the Anomaly Prediction (AP) task fundamentally relies on the existence of anomaly precursors, which are  gradual, progressive changes that signal an impending anomaly. Anomalies without precursors (e.g., instantaneous failures) are considered unpredictable. Therefore, we analyze AD datasets and select those suitable for AP evaluation. Our final selection includes MSL, PSM, SMD, GECCO, CICIDS, and Genesis, as detailed in Table \ref{tab:data}.

\subsection{Baselines and Evaluation Metrics}
Dedicated models for time series anomaly prediction (AP) are scarce. Many existing works focus on identifying anomaly precursors rather than predicting the timing of future anomalies, making them unsuitable for direct comparison. 
Consequently, we select FCM\cite{zhao2024fcm} and A2P\cite{park2025a2p}, which predict future anomalies by future context modeling, as our primary direct baseline. 
Additionally, to establish a more comprehensive performance benchmark, we also compare our model with various combinations of existing forecasting models and anomaly detection models.
For the forecasting models, we adopt five representative SOTA models: Linear-based DLinear and NLinear\cite{zeng2023dlinear}, Transformer-based PatchTST\cite{nie2022tst}, TimesNet\cite{wu2022timesnet}, and iTransformer\cite{liu2023itransformer}.
Regarding anomaly detection models, we adopt two widely recognized models: Anomaly Transformer (AT)\cite{xu2022anomaly} and DCdetector (DC)\cite{yang2023dcdetector}.

We avoid the widely-used Point Adjustment (PA) strategy, as it relies on the flawed assumption that detecting a single point within an anomalous segment constitutes a successful detection for the entire segment, which is unreasonable and unfairly\cite{wang2023d3r}. Consequently, we adopt the recently proposed Affiliated-F1-score (Aff-F1)  as our primary performance metric. The Aff-F1 provides a more reasonable and robust evaluation.

\subsection{Experiment Settings}\label{sec:setting}
We set a uniform input length of $L=192$ for all models and conduct experiments with $H = \{32, 64, 96, 128\}$. 
By default, We set the patch size, stride and embedding dimension to 16, 8, 256, respectively.
In DFM, Transformer layers and MSP modules are set to 3 and 2 respectively. 
The coefficients for weighing each loss term are $\lambda_{time}=1$ and $\lambda_{freq}=0.2$ for the REM, $\lambda_{main}=0.5$, $\lambda_{msp}=0.5$, and $\lambda_{contra}=1$ for the DFM. 

For each baseline method, we strictly follow the hyperparameter configurations recommended in their original papers to ensure a comprehensive and fair assessment.

All experiments are conducted on a single NVIDIA A800 GPU with 40GB of memory to exclude platform-related impacts.

\begin{table*}
\centering
\caption{Ablation studies for RED-F. The table shows the average Affiliated-F1-Score Metrics (averaged over different prediction lengths $H$) on all test datasets after removing specific components from REM and DFM. Specifically, w/o Frequency Modeling refers to substituting frequency-domain modeling with time-domain modeling, w/o graph denotes substituting the channel correlation discovering mechanism with with fixed channel dependencies strategies. The best results are marked in \textbf{bold}.}
\label{tab:ablation}
\begin{threeparttable}
\begin{small} 
\begin{tabular*}{\textwidth}{cl|@{\extracolsep{\fill}}ccccccc}
\toprule
\multicolumn{2}{c|}{Variations} & MSL & PSM & SMD & GECCO & CICIDS & Genesis & avg\\\midrule
\multicolumn{1}{c|}{\multirow{7}{*}{\makecell{REM}}} & w/o Frequency Modeling & \scalebox{0.78}{0.601} & \scalebox{0.78}{0.679} & \scalebox{0.78}{0.667} & \scalebox{0.78}{0.701} & \scalebox{0.78}{0.568} & \scalebox{0.78}{0.703} & \scalebox{0.78}{0.653} \\
\multicolumn{1}{c|}{} & w/o Frequency Patching & \scalebox{0.78}{0.597} & \scalebox{0.78}{0.711} & \scalebox{0.78}{0.681} & \scalebox{0.78}{0.764} & \scalebox{0.78}{0.619} & \scalebox{0.78}{0.753} & \scalebox{0.78}{0.687} \\
\multicolumn{1}{c|}{} & w/o Inter Attention & \scalebox{0.78}{0.664} & \scalebox{0.78}{0.733} & \scalebox{0.78}{0.700} & \scalebox{0.78}{0.739} & \scalebox{0.78}{0.620} & \scalebox{0.78}{0.762} & \scalebox{0.78}{0.703} \\
\multicolumn{1}{c|}{} & w/o Intra Attention & \scalebox{0.78}{0.674} & \scalebox{0.78}{0.734} & \scalebox{0.78}{0.685} & \scalebox{0.78}{0.747} & \scalebox{0.78}{0.632} & \scalebox{0.78}{0.779} & \scalebox{0.78}{0.708} \\
\multicolumn{1}{c|}{} & w/o Graph & \scalebox{0.78}{0.665} & \scalebox{0.78}{0.737} & \scalebox{0.78}{0.690} & \scalebox{0.78}{0.753} & \scalebox{0.78}{0.592} & \scalebox{0.78}{0.765} & \scalebox{0.78}{0.699} \\
\multicolumn{1}{c|}{} & w/o $\mathcal{L}_{freq}$ & \scalebox{0.78}{0.628} & \scalebox{0.78}{0.669} & \scalebox{0.78}{0.707} & \scalebox{0.78}{0.758} & \scalebox{0.78}{0.582} & \scalebox{0.78}{0.770} & \scalebox{0.78}{0.686} \\
\multicolumn{1}{c|}{} & w/o $\mathcal{L}_{time}$ & \scalebox{0.78}{0.609} & \scalebox{0.78}{0.634} & \scalebox{0.78}{0.664} & \scalebox{0.78}{0.770} & \scalebox{0.78}{0.596} & \scalebox{0.78}{0.727} & \scalebox{0.78}{0.667}
\\\midrule
\multicolumn{1}{c|}{\multirow{3}{*}{\makecell{DFM}}} & w/o Multi-Series Prediction& \scalebox{0.78}{0.671} & \scalebox{0.78}{0.696} & \scalebox{0.78}{0.691} & \scalebox{0.78}{0.757} & \scalebox{0.78}{0.609} & \scalebox{0.78}{0.744} & \scalebox{0.78}{0.695} \\
\multicolumn{1}{c|}{}& w/o $\mathcal{L}_{main}$ & \scalebox{0.78}{0.638} & \scalebox{0.78}{0.665} & \scalebox{0.78}{0.682} & \scalebox{0.78}{0.769} & \scalebox{0.78}{0.586} & \scalebox{0.78}{0.767} & \scalebox{0.78}{0.685} \\
\multicolumn{1}{c|}{} & w/o $\mathcal{L}_{contra}$ & \scalebox{0.78}{0.664} & \scalebox{0.78}{0.671} & \scalebox{0.78}{0.692} & \scalebox{0.78}{0.729} & \scalebox{0.78}{0.567} & \scalebox{0.78}{0.771} & \scalebox{0.78}{0.682} \\
\midrule
\multicolumn{2}{c|}{\textbf{{RED-F(ours)}}} & \textbf{\scalebox{0.78}{0.682}} & \textbf{\scalebox{0.78}{0.742}} & \textbf{\scalebox{0.78}{0.720}} & \textbf{\scalebox{0.78}{0.780}} & \textbf{\scalebox{0.78}{0.643}} & \textbf{\scalebox{0.78}{0.817}} & \textbf{\scalebox{0.78}{0.731}}\\\bottomrule
\end{tabular*}
\end{small} 
\end{threeparttable}
\end{table*}

\begin{figure*}[!htbp]
    \centering
    \subfigure[Patch Size]{\includegraphics[width=0.3\linewidth]{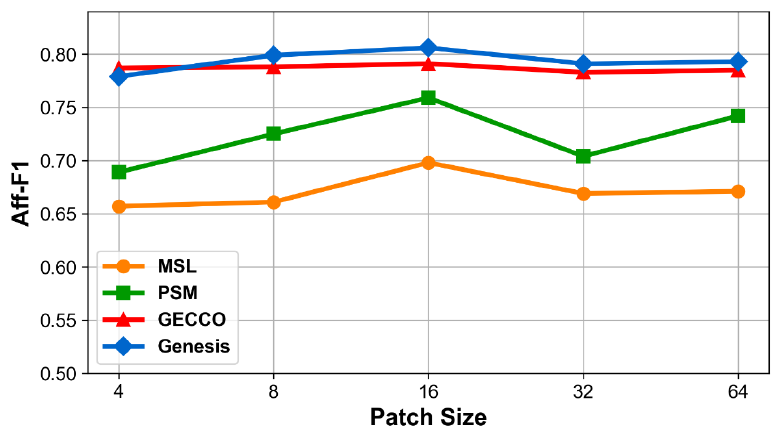}\label{fig:sen1}}
    \hfill
    \subfigure[MSP Num]{\includegraphics[width=0.3\linewidth]{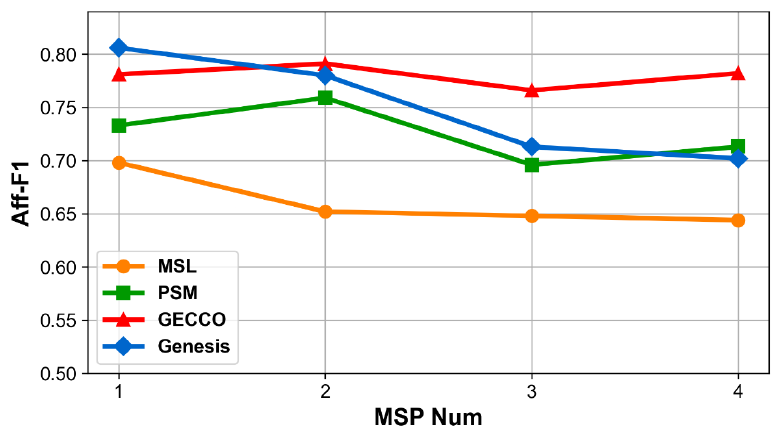}\label{fig:sen2}}
    \hfill
    \subfigure[$\lambda_{msp}$]{\includegraphics[width=0.3\linewidth]{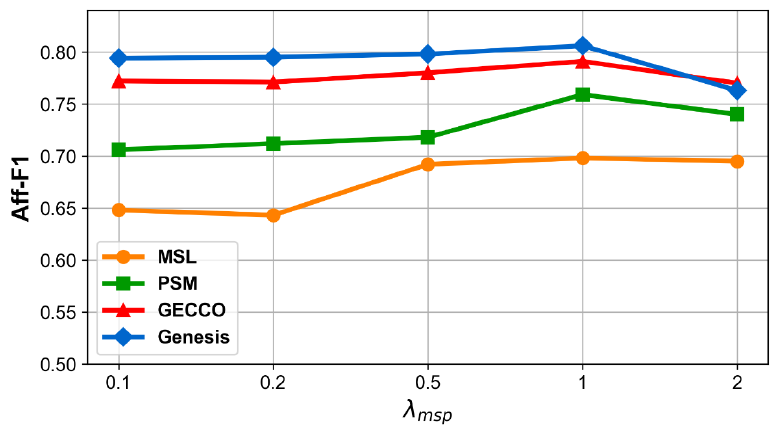}\label{fig:sen3}}
    \caption{Parameter sensitivity studies of RED-F, with input length $L$=196 and prediction horizon $H=$32.}
    \label{fig:sensitivity}
\end{figure*}

\begin{table*}[tbp]
\caption{Metric comparison of each model on the anomaly detection task. Results are shown for window length $L=100$, The Aff-P, Aff-R, and Aff-F1 scores for RED-F and the baseline models are evaluated on the six datasets. The best results are marked in \textbf{bold}.}
\label{tab:ad_main_results}
    \vspace{-5pt}
    \renewcommand{\arraystretch}{1.25} 
    \vskip 0.05in
    \centering
    \begin{small}
    \renewcommand{\multirowsetup}{\centering}
    \setlength{\tabcolsep}{4.35pt}
    \begin{tabular}{c|ccc|ccc|ccc|ccc|ccc|ccc}
    \toprule
    \multicolumn{1}{c}{\scalebox{1.05}Dataset} & \multicolumn{3}{c}{\scalebox{1}{MSL}} & \multicolumn{3}{c}{\scalebox{1}{PSM}} & \multicolumn{3}{c}{\scalebox{1}{SMD}} & \multicolumn{3}{c}{\scalebox{1}{Genesis}} & \multicolumn{3}{c}{\scalebox{1}{GECCO}} & \multicolumn{3}{c}{\scalebox{1}{CICIDS}}\\
    \cmidrule(lr){2-4}\cmidrule(lr){5-7}\cmidrule(lr){8-10}\cmidrule(lr){11-13}\cmidrule(lr){14-16}\cmidrule(lr){17-19}
    \multicolumn{1}{c}{\scalebox{1}{Metric}} &
    \scalebox{0.78}{Aff-P} & \scalebox{0.78}{Aff-R} & \scalebox{0.78}{Aff-F1} &\scalebox{0.78}{Aff-P} & \scalebox{0.78}{Aff-R} & \scalebox{0.78}{Aff-F1} &\scalebox{0.78}{Aff-P} & \scalebox{0.78}{Aff-R} & \scalebox{0.78}{Aff-F1} &\scalebox{0.78}{Aff-P} & \scalebox{0.78}{Aff-R} & \scalebox{0.78}{Aff-F1} &\scalebox{0.78}{Aff-P} & \scalebox{0.78}{Aff-R} & \scalebox{0.78}{Aff-F1} &\scalebox{0.78}{Aff-P} & \scalebox{0.78}{Aff-R} & \scalebox{0.78}{Aff-F1} \\

    \midrule
     \scalebox{0.9}{HBOS} & \scalebox{0.78}{0.520} & \scalebox{0.78}{0.982} & \scalebox{0.78}{0.680} & \scalebox{0.78}{0.621} & \scalebox{0.78}{0.700} & \scalebox{0.78}{0.658} & \scalebox{0.78}{0.557} & \scalebox{0.78}{0.722} & \scalebox{0.78}{0.629} & \scalebox{0.78}{0.564}  & \scalebox{0.78}{1.000} & \scalebox{0.78}{0.721} & \scalebox{0.78}{0.620} & \scalebox{0.78}{0.827}  & \scalebox{0.78}{0.708} & \scalebox{0.78}{0.538} & \scalebox{0.78}{0.545}  & \scalebox{0.78}{0.542}\\\rule{0pt}{4pt}
    
    \scalebox{0.9}{PCA} & \scalebox{0.78}{0.538} & \scalebox{0.78}{0.914} & \scalebox{0.78}{0.678} & \scalebox{0.78}{0.712} & \scalebox{0.78}{0.692} & \scalebox{0.78}{0.702} & \scalebox{0.78}{0.680} & \scalebox{0.78}{0.807} & \scalebox{0.78}{0.738} & \scalebox{0.78}{0.691}  & \scalebox{0.78}{0.991} & \scalebox{0.78}{0.814} & \scalebox{0.78}{0.646} & \scalebox{0.78}{1.000}  & \scalebox{0.78}{0.785} & \scalebox{0.78}{0.541} & \scalebox{0.78}{0.724}  & \scalebox{0.78}{0.619}\\\rule{0pt}{4pt}
          
    \scalebox{0.9}{IF} & \scalebox{0.78}{0.502} & \scalebox{0.78}{0.697} & \scalebox{0.78}{0.584} & \scalebox{0.78}{0.904} & \scalebox{0.78}{0.472} & \scalebox{0.78}{0.620} & \scalebox{0.78}{0.801} & \scalebox{0.78}{0.513} & \scalebox{0.78}{0.626} & \scalebox{0.78}{0.673}  & \scalebox{0.78}{0.951} & \scalebox{0.78}{0.788} & \scalebox{0.78}{0.647} & \scalebox{0.78}{0.315}  & \scalebox{0.78}{0.424} & \scalebox{0.78}{0.548} & \scalebox{0.78}{0.672}  & \scalebox{0.78}{0.604}\\\rule{0pt}{4pt}
    
    \scalebox{0.9}{Ocsvm} & \scalebox{0.78}{0.497} & \scalebox{0.78}{0.902} & \scalebox{0.78}{0.641} & \scalebox{0.78}{0.652} & \scalebox{0.78}{0.447} & \scalebox{0.78}{0.531} & \scalebox{0.78}{0.649} & \scalebox{0.78}{0.866} & \scalebox{0.78}{0.742} & \scalebox{0.78}{0.512}  & \scalebox{0.78}{1.000} & \scalebox{0.78}{0.677} & \scalebox{0.78}{0.499} & \scalebox{0.78}{1.000}  & \scalebox{0.78}{0.666} & \scalebox{0.78}{0.530} & \scalebox{0.78}{1.000}  & \scalebox{0.78}{0.693}\\\rule{0pt}{4pt}

    \scalebox{0.9}{TFAD} & \scalebox{0.78}{0.516} & \scalebox{0.78}{0.936} & \scalebox{0.78}{0.665} & \scalebox{0.78}{0.543} & \scalebox{0.78}{0.744} & \scalebox{0.78}{0.628} & \scalebox{0.78}{0.517} & \scalebox{0.78}{0.913} & \scalebox{0.78}{0.660} & \scalebox{0.78}{0.437}  & \scalebox{0.78}{0.687} & \scalebox{0.78}{0.535} & \scalebox{0.78}{0.526} & \scalebox{0.78}{0.776}  & \scalebox{0.78}{0.66} & \scalebox{0.78}{0.550} & \scalebox{0.78}{0.611}  & \scalebox{0.78}{0.579}\\\rule{0pt}{4pt}
    
    \scalebox{0.9}{AE} & \scalebox{0.78}{0.521} & \scalebox{0.78}{0.781} & \scalebox{0.78}{0.625} & \scalebox{0.78}{0.776} & \scalebox{0.78}{0.649} & \scalebox{0.78}{0.707} & \scalebox{0.78}{0.889} & \scalebox{0.78}{0.291} & \scalebox{0.78}{0.439} & \scalebox{0.78}{0.759}  & \scalebox{0.78}{0.976} & \scalebox{0.78}{0.854} & \scalebox{0.78}{0.836} & \scalebox{0.78}{0.810}  & \scalebox{0.78}{0.823} & \scalebox{0.78}{0.543} & \scalebox{0.78}{0.156}  & \scalebox{0.78}{0.243}\\\rule{0pt}{4pt}
    
    \scalebox{0.9}{DC} & \scalebox{0.78}{0.576} & \scalebox{0.78}{0.874} & \scalebox{0.78}{0.694} & \scalebox{0.78}{0.538} & \scalebox{0.78}{0.932} & \scalebox{0.78}{0.682} & \scalebox{0.78}{0.510} & \scalebox{0.78}{0.998} & \scalebox{0.78}{0.675} & \scalebox{0.78}{0.659}  & \scalebox{0.78}{0.943} & \scalebox{0.78}{0.776} & \scalebox{0.78}{0.567} & \scalebox{0.78}{0.872}  & \scalebox{0.78}{0.687} & \scalebox{0.78}{0.533} & \scalebox{0.78}{0.882}  & \scalebox{0.78}{0.664}\\\rule{0pt}{4pt}

    \scalebox{0.9}{ATrans} & \scalebox{0.78}{0.549} & \scalebox{0.78}{0.933} & \scalebox{0.78}{0.692} & \scalebox{0.78}{0.600} & \scalebox{0.78}{0.871} & \scalebox{0.78}{0.710} & \scalebox{0.78}{0.607} & \scalebox{0.78}{0.895} & \scalebox{0.78}{0.724} & \scalebox{0.78}{0.749}  & \scalebox{0.78}{1.000} & \scalebox{0.78}{0.856} & \scalebox{0.78}{0.690} & \scalebox{0.78}{0.903}  & \scalebox{0.78}{0.782} & \scalebox{0.78}{0.531} & \scalebox{0.78}{0.882}  & \scalebox{0.78}{0.560}\\\rule{0pt}{4pt}
    
    \scalebox{0.9}{DualTF} & \scalebox{0.78}{0.562} & \scalebox{0.78}{0.618} & \scalebox{0.78}{0.588} & \scalebox{0.78}{0.622} & \scalebox{0.78}{0.868} & \scalebox{0.78}{0.725} & \scalebox{0.78}{0.527} & \scalebox{0.78}{0.956} & \scalebox{0.78}{0.679} & \scalebox{0.78}{0.683}  & \scalebox{0.78}{0.996} & \scalebox{0.78}{0.810} & \scalebox{0.78}{0.633} & \scalebox{0.78}{0.786}  & \scalebox{0.78}{0.701} & \scalebox{0.78}{0.553} & \scalebox{0.78}{0.926}  & \scalebox{0.78}{0.692}\\\rule{0pt}{4pt}

    \scalebox{0.9}{Modern} & \scalebox{0.78}{0.578} & \scalebox{0.78}{0.975} & \scalebox{0.78}{0.723} & \scalebox{0.78}{0.734} & \scalebox{0.78}{0.941} & \scalebox{0.78}{0.825} & \scalebox{0.78}{0.755} & \scalebox{0.78}{0.948} & \scalebox{0.78}{0.840} & \scalebox{0.78}{0.728}  & \scalebox{0.78}{0.974} & \scalebox{0.78}{0.833} & \scalebox{0.78}{0.808} & \scalebox{0.78}{0.998}  & \scalebox{0.78}{0.893} & \scalebox{0.78}{0.563} & \scalebox{0.78}{0.781}  & \scalebox{0.78}{0.654}\\\rule{0pt}{4pt}

    \scalebox{0.9}{CATCH} & \scalebox{0.78}{0.579} & \scalebox{0.78}{0.945} & \scalebox{0.78}{0.718} & \scalebox{0.78}{0.757} & \scalebox{0.78}{0.929} & \scalebox{0.78}{0.834} & \scalebox{0.78}{0.764} & \scalebox{0.78}{0.929} & \scalebox{0.78}{0.839} & \scalebox{0.78}{0.748}  & \scalebox{0.78}{0.952} & \scalebox{0.78}{0.838} & \scalebox{0.78}{0.811} & \scalebox{0.78}{0.998}  & \scalebox{0.78}{0.895} & \scalebox{0.78}{0.594} & \scalebox{0.78}{0.842}  & \scalebox{0.78}{0.700}\\\rule{0pt}{4pt}
    
    \scalebox{0.9}{CAROTS} & \scalebox{0.78}{0.523} & \scalebox{0.78}{0.998} & \scalebox{0.78}{0.686} & \scalebox{0.78}{0.626} & \scalebox{0.78}{0.978} & \scalebox{0.78}{0.763} & \scalebox{0.78}{0.726} & \scalebox{0.78}{0.762} & \scalebox{0.78}{0.744} & \scalebox{0.78}{0.611}  & \scalebox{0.78}{0.937} & \scalebox{0.78}{0.740} & \scalebox{0.78}{0.703} & \scalebox{0.78}{0.807}  & \scalebox{0.78}{0.751} & \scalebox{0.78}{0.600} & \scalebox{0.78}{1.000}  & \scalebox{0.78}{0.750}\\\rule{0pt}{4pt}
    \scalebox{0.9}{MtsCID} & \scalebox{0.78}{0.516} & \scalebox{0.78}{0.960} & \scalebox{0.78}{0.672} & \scalebox{0.78}{0.543} & \scalebox{0.78}{0.839} & \scalebox{0.78}{0.659} & \scalebox{0.78}{0.589} & \scalebox{0.78}{0.907} & \scalebox{0.78}{0.714} & \scalebox{0.78}{0.506}  & \scalebox{0.78}{0.888} & \scalebox{0.78}{0.644} & \scalebox{0.78}{0.598} & \scalebox{0.78}{0.951}  & \scalebox{0.78}{0.734} & \scalebox{0.78}{0.514} & \scalebox{0.78}{0.258}  & \scalebox{0.78}{0.344}\\
    \midrule
    {\scalebox{1.0}{\textbf{\MakeUppercase{RED-F}}}} & \scalebox{0.78}{0.634} & \scalebox{0.78}{0.893} & \boldres{\scalebox{0.78}{0.742}} & \scalebox{0.78}{0.750} & \scalebox{0.78}{0.959} & \boldres{\scalebox{0.78}{0.842}} & \scalebox{0.78}{0.768} & \scalebox{0.78}{0.933} & \boldres{\scalebox{0.78}{0.842}} & \scalebox{0.78}{0.818} &\scalebox{0.78}{0.957} & \boldres{\scalebox{0.78}{0.882}} & \scalebox{0.78}{0.820} & \scalebox{0.78}{0.998} & \boldres{\scalebox{0.78}{0.900}} & \scalebox{0.78}{0.646} & \scalebox{0.78}{0.957} & \boldres{\scalebox{0.78}{0.771}} \\
    \bottomrule
    \end{tabular}
    \end{small}
    \vspace{-10pt}
\end{table*}

\begin{table*}[!htbp]
    \centering
    \caption{Results of the TSF task on seven real-world datasets. Results are averaged over input $L=96$ and prediction horizons $H = \{96, 192, 336, 720\}$. Metrics are Mean Square Error (MSE) and Mean Absolute Error (MAE). Lower values are better. The best results are marked in \textbf{bold}.}
    \label{tab:tsf_main_results}
    \vspace{2pt}
    \renewcommand{\arraystretch}{1.25}
    \setlength{\tabcolsep}{7.45pt}

    \begin{tabular}{l|cc|cc|cc|cc|cc|cc|cc}
        \toprule
        \scalebox{1.05}{Dataset}& 
        \multicolumn{2}{c}{\scalebox{0.9}{\textbf{RED-F}}} &
        \multicolumn{2}{c}{\scalebox{0.9}{iTransformer}} &
        \multicolumn{2}{c}{\scalebox{0.9}{RLinear}} &
        \multicolumn{2}{c}{\scalebox{0.9}{PatchTST}} &
        \multicolumn{2}{c}{\scalebox{0.9}{Autoformer}} &
        \multicolumn{2}{c}{\scalebox{0.9}{FEDformer}} &
        \multicolumn{2}{c}{\scalebox{0.9}{DLinear}} \\
        \cmidrule(lr){2-3} 
        \cmidrule(lr){4-5} 
        \cmidrule(lr){6-7} 
        \cmidrule(lr){8-9} 
        \cmidrule(lr){10-11} 
        \cmidrule(lr){12-13} 
        \cmidrule(lr){14-15}
        \scalebox{1}{Metric} & \scalebox{0.78}{MSE} & \scalebox{0.78}{MAE} & \scalebox{0.78}{MSE} & \scalebox{0.78}{MAE} & \scalebox{0.78}{MSE} & \scalebox{0.78}{MAE} & \scalebox{0.78}{MSE} & \scalebox{0.78}{MAE} & \scalebox{0.78}{MSE} & \scalebox{0.78}{MAE} & \scalebox{0.78}{MSE} & \scalebox{0.78}{MAE} & \scalebox{0.78}{MSE} & \scalebox{0.78}{MAE}\\
        \midrule
        ETTh1 & \scalebox{0.78}{0.442} & \scalebox{0.78}{0.435} & \scalebox{0.78}{0.454} & \scalebox{0.78}{0.447} & \scalebox{0.78}{0.446} & \textbf{\scalebox{0.78}{0.434}} & \scalebox{0.78}{0.469} & \scalebox{0.78}{0.454} & \scalebox{0.78}{0.496} & \scalebox{0.78}{0.487} & \textbf{\scalebox{0.78}{0.440}} & \scalebox{0.78}{0.460} & \scalebox{0.78}{0.456} & \scalebox{0.78}{0.452} \\
        ETTh2 & \scalebox{0.78}{0.387} & \scalebox{0.78}{0.408} & \scalebox{0.78}{0.383} & \scalebox{0.78}{0.407} & \textbf{\scalebox{0.78}{0.374}} & \textbf{\scalebox{0.78}{0.398}} & \scalebox{0.78}{0.387} & \scalebox{0.78}{0.407} & \scalebox{0.78}{0.450} & \scalebox{0.78}{0.459} & \scalebox{0.78}{0.437} & \scalebox{0.78}{0.449} & \scalebox{0.78}{0.559} & \scalebox{0.78}{0.515} \\
        ETTm1 & \textbf{\scalebox{0.78}{0.380}} & \textbf{\scalebox{0.78}{0.398}} & \scalebox{0.78}{0.407} & \scalebox{0.78}{0.410} & \scalebox{0.78}{0.414} & \scalebox{0.78}{0.407} & \scalebox{0.78}{0.387} & \scalebox{0.78}{0.400} & \scalebox{0.78}{0.588} & \scalebox{0.78}{0.517} & \scalebox{0.78}{0.448} & \scalebox{0.78}{0.452} & \scalebox{0.78}{0.403} & \scalebox{0.78}{0.407} \\
        ETTm2 & \textbf{\scalebox{0.78}{0.278}} & \textbf{\scalebox{0.78}{0.325}} & \scalebox{0.78}{0.288} & \scalebox{0.78}{0.332} & \scalebox{0.78}{0.286} & \scalebox{0.78}{0.327} & \scalebox{0.78}{0.281} & \scalebox{0.78}{0.326} & \scalebox{0.78}{0.327} & \scalebox{0.78}{0.371} & \scalebox{0.78}{0.305} & \scalebox{0.78}{0.349} & \scalebox{0.78}{0.350} & \scalebox{0.78}{0.401} \\
        Weather & \textbf{\scalebox{0.78}{0.243}} & \textbf{\scalebox{0.78}{0.274}} & \scalebox{0.78}{0.258} & \scalebox{0.78}{0.278} & \scalebox{0.78}{0.272} & \scalebox{0.78}{0.291} & \scalebox{0.78}{0.259} & \scalebox{0.78}{0.281} & \scalebox{0.78}{0.338} & \scalebox{0.78}{0.382} & \scalebox{0.78}{0.309} & \scalebox{0.78}{0.360} & \scalebox{0.78}{0.265} & \scalebox{0.78}{0.317} \\
        Electricity & \textbf{\scalebox{0.78}{0.178}} & \textbf{\scalebox{0.78}{0.270}} & \textbf{\scalebox{0.78}{0.178}} & \textbf{\scalebox{0.78}{0.270}} & \scalebox{0.78}{0.219} & \scalebox{0.78}{0.298} & \scalebox{0.78}{0.205} & \scalebox{0.78}{0.290} & \scalebox{0.78}{0.227} & \scalebox{0.78}{0.338} & \scalebox{0.78}{0.214} & \scalebox{0.78}{0.327} & \scalebox{0.78}{0.212} & \scalebox{0.78}{0.300} \\
        Solar & \textbf{\scalebox{0.78}{0.225}} & \scalebox{0.78}{0.267} & \scalebox{0.78}{0.233} & \textbf{\scalebox{0.78}{0.262}} & \scalebox{0.78}{0.369} & \scalebox{0.78}{0.356} & \scalebox{0.78}{0.270} & \scalebox{0.78}{0.307} & \scalebox{0.78}{0.885} & \scalebox{0.78}{0.711} & \scalebox{0.78}{0.291} & \scalebox{0.78}{0.381} & \scalebox{0.78}{0.330} & \scalebox{0.78}{0.401} \\

        \bottomrule
    \end{tabular}
\end{table*}

\subsection{Main Results}
The results of the Anomaly Prediction experiments are demonstrated in Table \ref{tab:main-result}. It can be seen that our proposed RED-F outperforms all baselines under the widely used Affiliated-F1-score Metric. Such improvement of RED-F can be attributed to three aspects: (a) The frequency-domain modeling can better capture the gradual pattern changes of subtle anomaly precursors; (b) A new training objective with MSP strengthens the sensitivity of forecasting models to anomaly precursors; (c) Dual-stream contrastive forecasting further amplifies the predicted anomalous signals.

Among all the datasets, RED-F achieves the most improvement on PSM and CICIDS, which contain a large number of variables and exhibit more pronounced correlations between variables. In particular, the improvement of Aff-F1 comes to 10.10\% and 4.30\% on PSM, and 14.00\% and 14.40\% on CICIDS, compared to the methods A2P and FCM. Such improvement of RED-F can be attributed to the design of REM, which models complex correlations between variables using the frequency-domain weighted distance method. 
Note that RED-F are robust in datasets from various prediction horizons, which implies that our method is robust to the long-term anomaly prediction tasks, which is important for real-world applications.


\subsection{Ablation Study}
To evaluate the impact of different modules within RED-F, we conduct ablation studies focusing on the following components,
for REM: (1) Substitute frequency-domain modeling with time-domain modeling. (2) Remove the frequency patching operation. (3) Delete one of the dual-perspective modeling of frequency-domain separately. (4) Substitute the frequency-domain weighted distance method that discovers channel correlations with fixed Channel Strategies (Channel Dependencies). (5) Delete one of the two optimization objectives ($\mathcal{L}_{time}$ and $\mathcal{L}_{freq}$) separately. For DFM: (1) Remove Multi-Series Prediction. (2) Delete one of the two optimization objectives ($\mathcal{L}_{main}$ and $\mathcal{L}_{contra}$) separately.


As shown in Table \ref{tab:ablation}, removing any component leads to a significant performance degradation, which verifies that each component of our design is effective and necessary.

The analysis of REM confirm our strategy for handling weak precursors.
Substituting frequency-domain modeling with time-domain modeling results in the most significant performance degradation, which verifies the effectiveness of frequency-domain modeling in identifying the gradual pattern changes associated with anomaly precursors.
When removing frequency patching, or when deleting one of the dual-perspective modeling, the model performance decreases in both cases. This indicates the need for fine-grained frequency modeling to capture weak anomaly precursors. 
Also, substituting the weighted-distance graph with a Channel Dependency strategy degrades the model performance, demonstrating the effectiveness of the channel correlation discovering method. 
Besides, removing $\mathcal{L}_{time}$ or $\mathcal{L}_{freq}$ leads to a decline in model performance. This fully indicates that time-frequency synergy is essential for modeling diverse patterns of anomaly precursors.

As for the analysis of DFM, the performance decline from removing  MSP confirms that a new training objective with MSP is necessary for forecasting models faced with subtle anomaly precursors. Besides, the balance of $\mathcal{L}_{main}$ and $\mathcal{L}_{contra}$ is also crucial. Deleting $\mathcal{L}_{main}$ guides the model to learn a trivial solution (e.g., making both predictions identical), leading to many false negatives.  Conversely, removing $\mathcal{L}_{contra}$ fails to constrain the dual-stream difference, resulting in many false positives.


\subsection{Analysis of REM for Anomaly Detection}
To investigate the effectiveness of REM for Anomaly Detection, we employ REM as a reconstruction-based model for anomaly detection tasks. REM is compared against representative deep learning (MtsCID\cite{xie2025mtscid}, CAROTS\cite{kim2025carots}, CATCH\cite{wu2025catch}, ModernTCN\cite{luo2024moderntcn}, Anomaly Transformer (ATrans)\cite{xu2022anomaly}, DCdetector (DC)\cite{yang2023dcdetector}, AutoEncoder (AE)\cite{sakurada2014ae}, TFAD\cite{zhang2022tfad}) and traditional machine learning (OCSVM\cite{scholkopf1999support}, IF\cite{liu2008isolation}, PCA\cite{shyu2003novel}, HBOS\cite{goldstein2012histogram}) baselines. All methods use a uniform window length of $L=100$. We replicate and run MtsCID, CAROTS, and CATCH using the publicly available optimal parameters. For each baseline models, we adhere to their established implementations. We record the Affiliated-F1-score to measure the performance of different models, with detailed results shown in Table \ref{tab:ad_main_results}.


The results presented in Table \ref{tab:ad_main_results} demonstrate the superiority of REM in anomaly detection tasks. This indicates that the fine-grained frequency-domain modeling can not only identify anomaly precursors, but also help to capture real anomalies. We consider this frequency-domain modeling is good at capturing series anomalies, thus achieving competitive performance against state-of-the-art models.



\begin{table*}[htbp]
\centering
\caption{Comparison of FCM and A2P in terms of training time, test time, total parameters and Aff-F1 result. Training time refers to the time cost to train the model for 5 epochs with a fixed batch size of 128, while test time denotes the duration needed to process the entire test datasets. Total params indicates the total number of trainable parameters within the model}
\label{tab:efficiency_comparison}
\setlength{\tabcolsep}{17pt}
\renewcommand{\arraystretch}{1.25}
\begin{tabular}{l|lcccc}
\toprule
\textbf{Dataset} & \textbf{Method} & \textbf{Training Time (s)} & \textbf{Test Time (s)} & \textbf{Total Params (M)} & \textbf{Aff-F1} \\
\hline
\multirow{3}{*}{GECCO} & FCM & \scalebox{0.78}{255.23} & \scalebox{0.78}{25.37} & \scalebox{0.78}{4.11} & \scalebox{0.78}{0.676} \\
                       & A2P  & \scalebox{0.78}{201.79} & \scalebox{0.78}{38.62} & \scalebox{0.78}{4.40} & \scalebox{0.78}{0.660} \\
                       & \textbf{RED-F} & \textbf{\scalebox{0.78}{111.18}} & \textbf{\scalebox{0.78}{19.28}} & \textbf{\scalebox{0.78}{3.62}} & \textbf{\scalebox{0.78}{0.793}} \\
\hline
\multirow{3}{*}{CICIDS} & FCM & \scalebox{0.78}{489.09} & \scalebox{0.78}{20.66} & \textbf{\scalebox{0.78}{4.17}} & \scalebox{0.78}{0.567} \\
                       & A2P  & \scalebox{0.78}{323.94} & \textbf{\scalebox{0.78}{15.09}} & \scalebox{0.78}{4.53} & \scalebox{0.78}{0.565} \\
                       & \textbf{RED-F} & \textbf{\scalebox{0.78}{261.55}} & \scalebox{0.78}{21.88} & \scalebox{0.78}{5.03} & \textbf{\scalebox{0.78}{0.678}} \\
\bottomrule
\end{tabular}
\end{table*}

\begin{figure*}[t]
    \centering
    \subfigure[Normal pattern]{\includegraphics[width=0.48\linewidth]{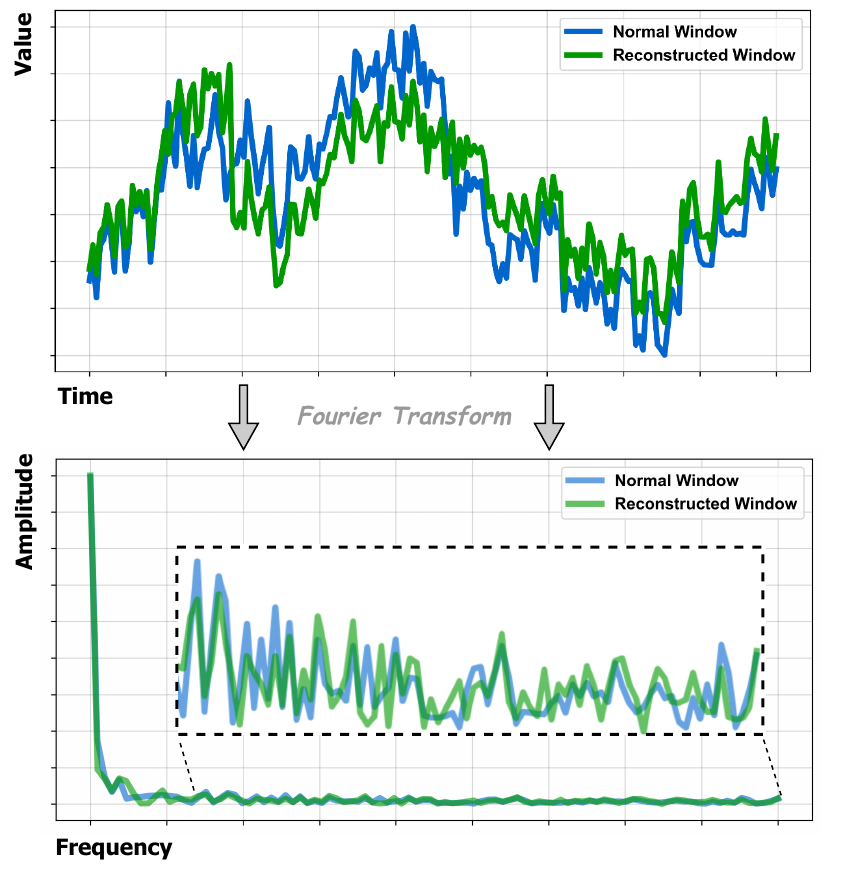}\label{fig:rem1}}
    \subfigure[Anomaly precursor]{\includegraphics[width=0.48\linewidth]{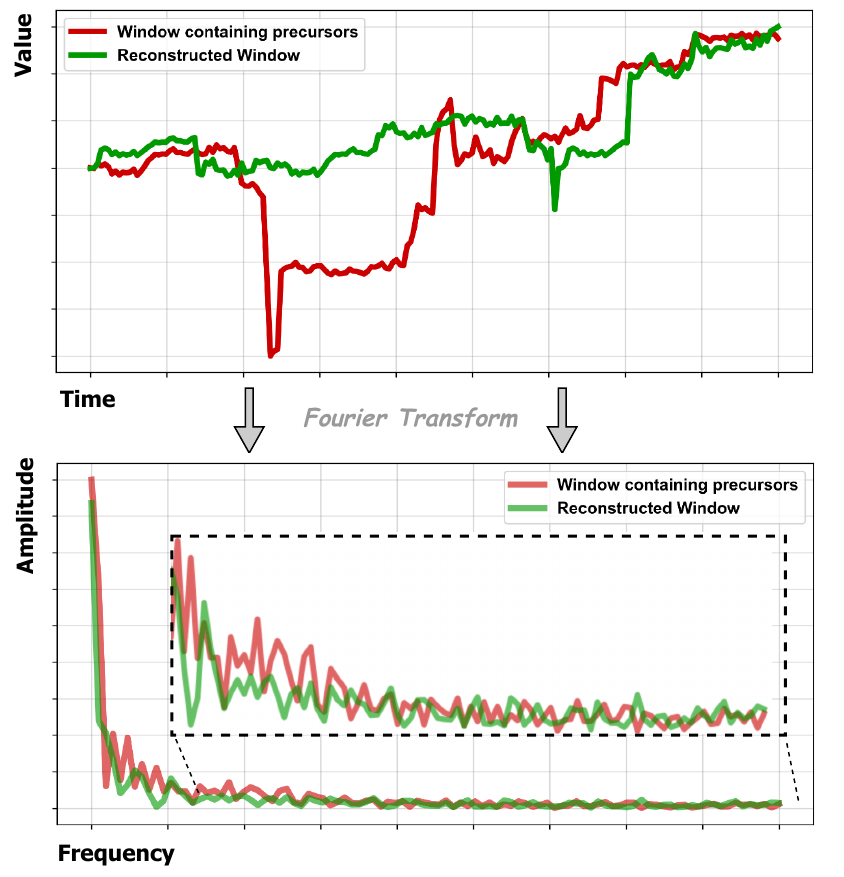}\label{fig:rem2}}
    \caption{The comparison between the reconstruction of a normal window and a window containing anomaly precursors. The normal window and its reconstructed result exhibit high similarity in the frequency domain, whereas the window containing precursors shows pronounced differences from its reconstruction in the frequency domain.}
    \label{fig:rem}
\end{figure*}

\subsection{Analysis of DFM for Long-term Forecasting}
We also evaluate the effectiveness of DFM in the time series long-term forecasting (TSF) tasks. For this task, DFM is separated into a single-stream forecasting model and compared against six advanced TSF baselines (e.g., iTransformer\cite{liu2023itransformer}, PatchTST\cite{nie2022tst}) on seven datasets: Solar, Weather, Electricity, and 4 ETT datasets (ETTh1,ETTh2,ETTm1,ETTm2). These datasets are extensively utilized, covering multiple fields including weather and energy management.

As shown in Table \ref{tab:tsf_main_results}, DFM achieve competitive performance compared to state-of-the-art models. This advantage is primarily attributed to the MSP modules, which enable the model to fully consider the impact of future context when predicting the current window, effectively enhancing its accuracy.

\subsection{Hyper-parameter Sensitivity Analysis}
We also study the parameter sensitivity of RED-F.
Figure \ref{fig:sen1} shows the performance under different patch size of the frequency domain information. Being too small (insufficient information) or too large (redundancy/noise) is suboptimal. For our primary evaluation, the patch size is usually set as 8 or 16.
Also, 
Figure \ref{fig:sen2} shows the impact of different numbers of MSP modules ($n$) on performance. The optimal number of MSP heads shows the dependency of the datasets. MSL/Genesis achieves optimal performance at $n=1$, while PSM/GECCO requires $n=2$.
Besides, we evaluate the performance under different $\lambda_{msp}$ on Figure \ref{fig:sen3}.
We find that $\lambda_{msp}$ is mostly stable and easy to tune in the range of 0.1 to 1, which shows good robustness and reduces the difficulty of hyperparameter tuning.

\begin{figure*}
    \centering
    \includegraphics[width=\linewidth]{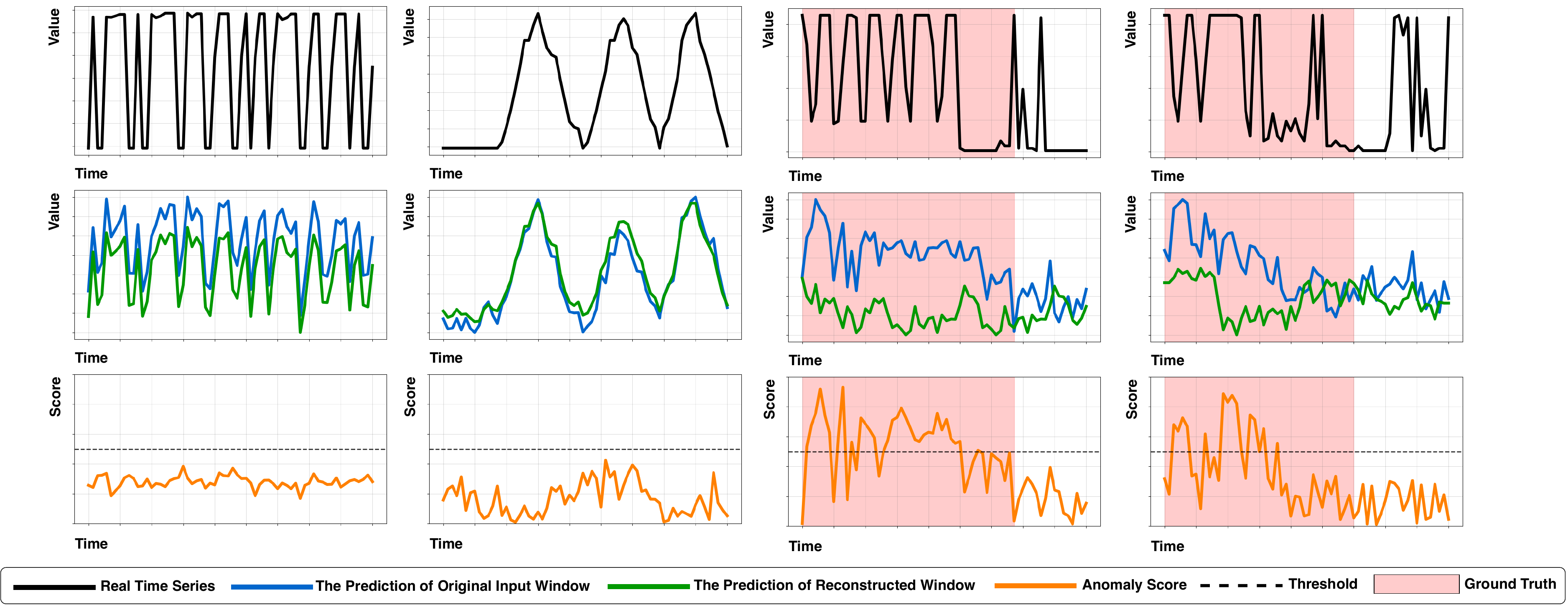}
    \caption{DFM prediction results on the SMD dataset. The red and blue curves are the predictions from the original and reconstruction streams, respectively. The green curve is the final anomaly score, and the pink background indicates the ground truth anomaly.}
    \label{fig:dfm}
\end{figure*}

\subsection{Analysis of model efficiency}
We provide a comparison of current time series anomaly prediction methods, including FCM and A2P, in terms of efficiency on the GECCO and CICIDS datasets. As shown in Table \ref{tab:efficiency_comparison}, RED-F achieves comparable test-time efficiency and parameters to FCM and A2P on the GECCO datasets, while its Aff-F1 significantly outperforms the other models. However, on datasets with a large number of channels (e.g., CICIDS), RED-F has longer test time and higher parameters than baselines. This is primarily due to the complex design of REM, which explicitly models intricate inter-channel dependencies. Consequently, as the number of channels increases, the computational efficiency of RED-F decreases. Considering performance, the differences in inference time and parameters between our model and other baselines are acceptable. 
We will address this issue in future work by designing more lightweight approaches to model inter-channel dependencies or by leveraging knowledge distillation to obtain a more efficient model.
\begin{figure}[t]
    \centering
\includegraphics[width=\linewidth]{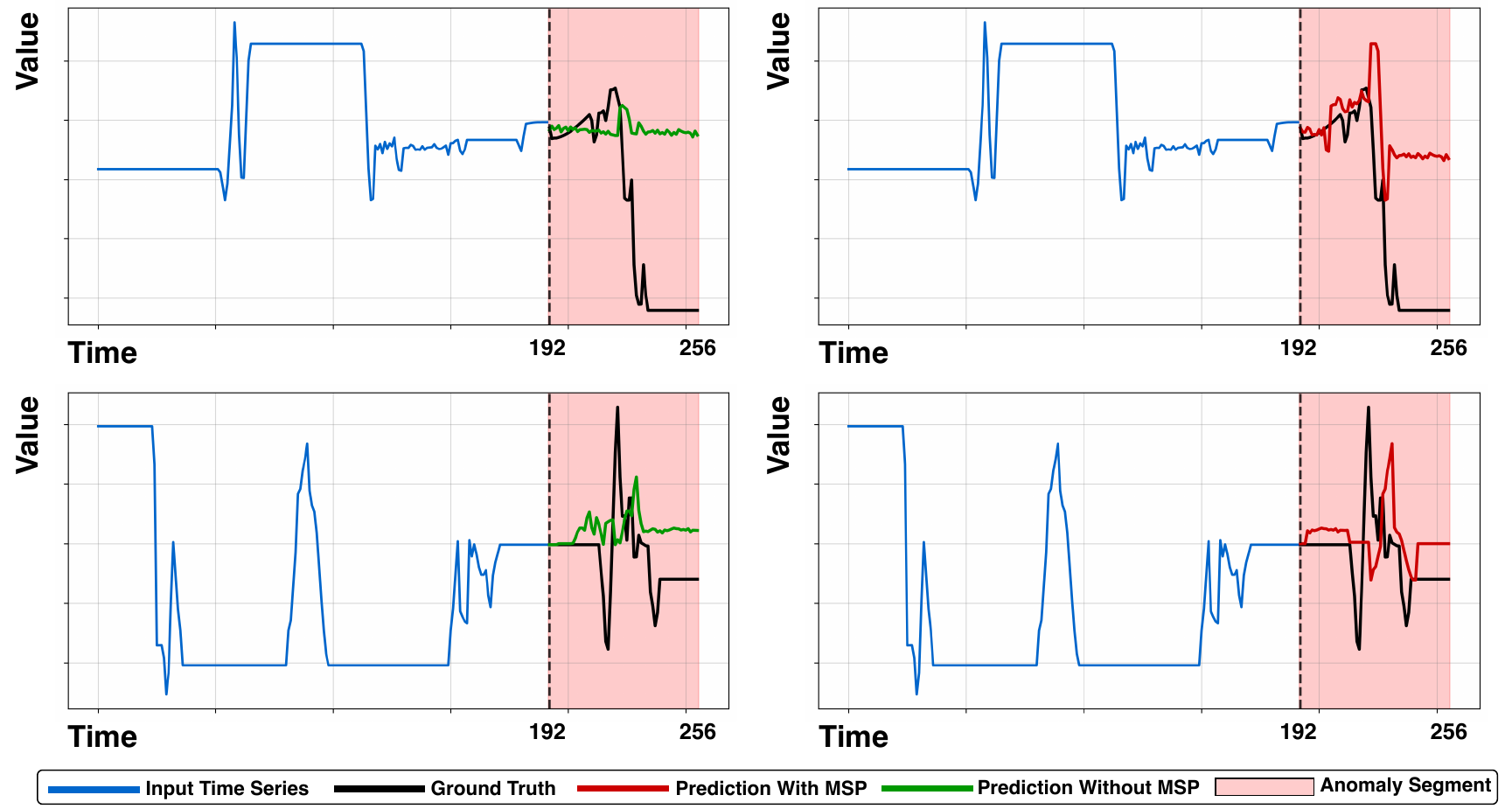}
 \caption{The comparison of the ground truth with predictions from the model trained with MSP and the model trained without MSP. These examples are from the SMD dataset, using $L=192$ and $H=32$. The MSP-trained model is visibly more sensitive to anomaly precursors, capturing gradual pattern changes earlier.}
 \label{fig:msp-exp}
\end{figure}
\subsection{Visualization}
\subsubsection{REM}
We conduct a comparative visualization analysis of normal pattern reconstruction and anomaly precursor reconstruction on SMD dataset to demonstrate the validity of REM. As shown in Figure \ref{fig:rem}, REM proactively captures the anomaly precursor and reverts it to the normal pattern. Figure \ref{fig:rem} also illustrates the frequency-domain differences between the original windows and their reconstructions. The normal window and its reconstruction exhibit highly similar frequency characteristics, whereas the window containing anomaly precursors differs substantially from its reconstruction in the frequency domain. This observation verifies the effectiveness of frequency-domain modeling in capturing the gradual pattern changes associated with anomaly precursors.
\subsubsection{DFM} 
We visualize the contrastive forecasting results of RED-F to demonstrate the effectiveness of DFM. The results are shown in Figure \ref{fig:dfm}. For normal future windows, the predictions based on the original window and its reconstructed counterpart are highly consistent, leading to a low future anomaly score. In contrast, for anomalous future windows, the prediction derived from the reconstructed window, which is more aligned with normal patterns, exhibits a stable forecast. Meanwhile, the prediction based on the original window deviates markedly from this stable result, leading to high future anomaly scores. The reason the prediction based on the original window exhibits stronger deviations from the stable forecast stems from our proposed MSP training objective. We further illustrate its effectiveness in Figure \ref{fig:msp-exp}. The model without MSP fails to generate strong predictions in response to subtle anomaly precursors. Conversely, the model equipped with MSP demonstrates higher sensitivity to anomaly precursors, producing stronger forecasts that more closely track the actual anomalies.

\section{Conclusion}
In this paper, we propose RED-F to address the challenging problem of unsupervised multivariate time series anomaly prediction. 
To sum up, it employs the Reconstruction-Elimination Model to generate a stable normal baseline, and then utilizes the Dual-Stream Contrastive Forecasting Model
to amplify the predicted future anomalous signals. To strengthen the sensitivity of forecasting models to anomaly precursors, we design a Multi-Series Prediction objective to leverage future context.
This innovative framework transforms the difficult task of faint signal detection into a simpler and more robust relative comparison. Extensive experiments demonstrate that RED-F significantly outperforms all baselines, showing particular robustness in anomaly prediction tasks. Future work will be done to (a) explore more lightweight approaches to model inter-channel dependencies and  leverage knowledge distillation to obtain a more efficient model, (b) investigate the integration of our dual-stream contrastive forecasting strategy with other types of advanced forecasting backbones, such as state space models (SSMs). Additionally, we are going to extend RED-F to other challenging applications, such as Large Time Series Models and AIOps.

\end{document}